\pdfoutput=1

\documentclass[11pt]{article}

\usepackage[final]{acl}

\usepackage{times}
\usepackage{latexsym}

\usepackage[T1]{fontenc}

\usepackage[utf8]{inputenc}

\usepackage{microtype}

\usepackage{inconsolata}

\usepackage{graphicx}
\usepackage{subcaption}
\usepackage{multirow}
\usepackage{placeins}

\usepackage[table]{xcolor}
\usepackage{pgfplotstable}
\usepackage{colortbl}

\usepackage{booktabs}
\usepackage{listings}
\usepackage{algorithm}
\usepackage{algpseudocode}
\usepackage{stfloats}

\definecolor{lightgreen}{rgb}{0.8, 1, 0.8}
\definecolor{lightblue}{rgb}{0.8, 0.8, 1}
\definecolor{lightpink}{rgb}{1.0, 0.6, 0.8}
\definecolor{lightyellow}{rgb}{1.0, 1.0, 0.2}

\newcommand{\Krel}{$\mathcal{K}_{H,M}^{rel}$}
\newcommand{\Kmodel}{$\mathcal{K}_{C,L}^{model}$}
\newcommand{\Inc}{$\mathcal{I}_{H \subset M}$}
\newcommand{\Rllm}{$\mathcal{R}_{LLM}$}
\newcommand{\Rhum}{$\mathcal{R}_{Human}$}

\usepackage{caption}

\title{The Alchemy of Thought: \\
Understanding In-Context Learning Through Supervised Classification}

\author{Harshita Narnoli \and Mihai Surdeanu \\
        Department of Computer Science, University of Arizona, Tucson, AZ, USA \\
        \texttt{\{harshitanarnoli,msurdeanu\}@arizona.edu}}

\begin{document}
\maketitle
\begin{abstract}


In-context learning (ICL) has become a prominent paradigm to rapidly customize LLMs to new tasks without fine-tuning. However, despite the empirical evidence of its usefulness, we still do not truly understand how ICL works. In this paper, we compare the behavior of in-context learning with supervised classifiers trained on ICL demonstrations to investigate three research questions: (1) Do LLMs with ICL behave similarly to classifiers trained on the same examples? (2) If so, which classifiers are closer, those based on gradient descent (GD) or those based on k-nearest neighbors (kNN)? (3) When they do not behave similarly, what conditions are associated with differences in behavior? Using text classification as a use case, with six datasets and three LLMs, we observe that LLMs behave similarly to these classifiers when the relevance of demonstrations is high. On average, ICL is closer to kNN than logistic regression, giving empirical evidence that the attention mechanism behaves more similarly to kNN than GD. However, when demonstration relevance is low, LLMs perform better than these classifiers, likely because LLMs can back off to their parametric memory, a luxury these classifiers do not have.



\end{abstract}

\section{Introduction}

In-context learning (ICL) has emerged as the dominant paradigm for adapting large language models (LLMs) to new tasks without requiring any parameter updates. ICL achieves this by including a few relevant examples (or demonstrations) in the LLM's prompt \cite{NEURIPS2020_1457c0d6,wei2022emergentabilitieslargelanguage}. However, despite the considerable amount of empirical evidence for ICL's usefulness, the reasons behind its success remain unclear.
On one hand, a few notable works establish an equivalence between ICL and gradient descent (GD) based on synthetic settings \cite{pmlr-v202-von-oswald23a,vonoswald2024uncoveringmesaoptimizationalgorithmstransformers,akyürek2023learningalgorithmincontextlearning,ahn2023transformerslearnimplementpreconditioned}. 
These works hypothesize that transformers use internal optimization using gradient-based algorithms to learn in-context. On the other hand, \citet{deutch2024incontextlearninggradientdescent} observes a weak correlation between ICL and GD, uncovering several major discrepancies in the flow of information throughout the model between the two processes. These contradictions suggest that we still do not yet know {\em how} ICL works. 

\begin{figure}[t]
  \centering
  \includegraphics[width=\columnwidth]{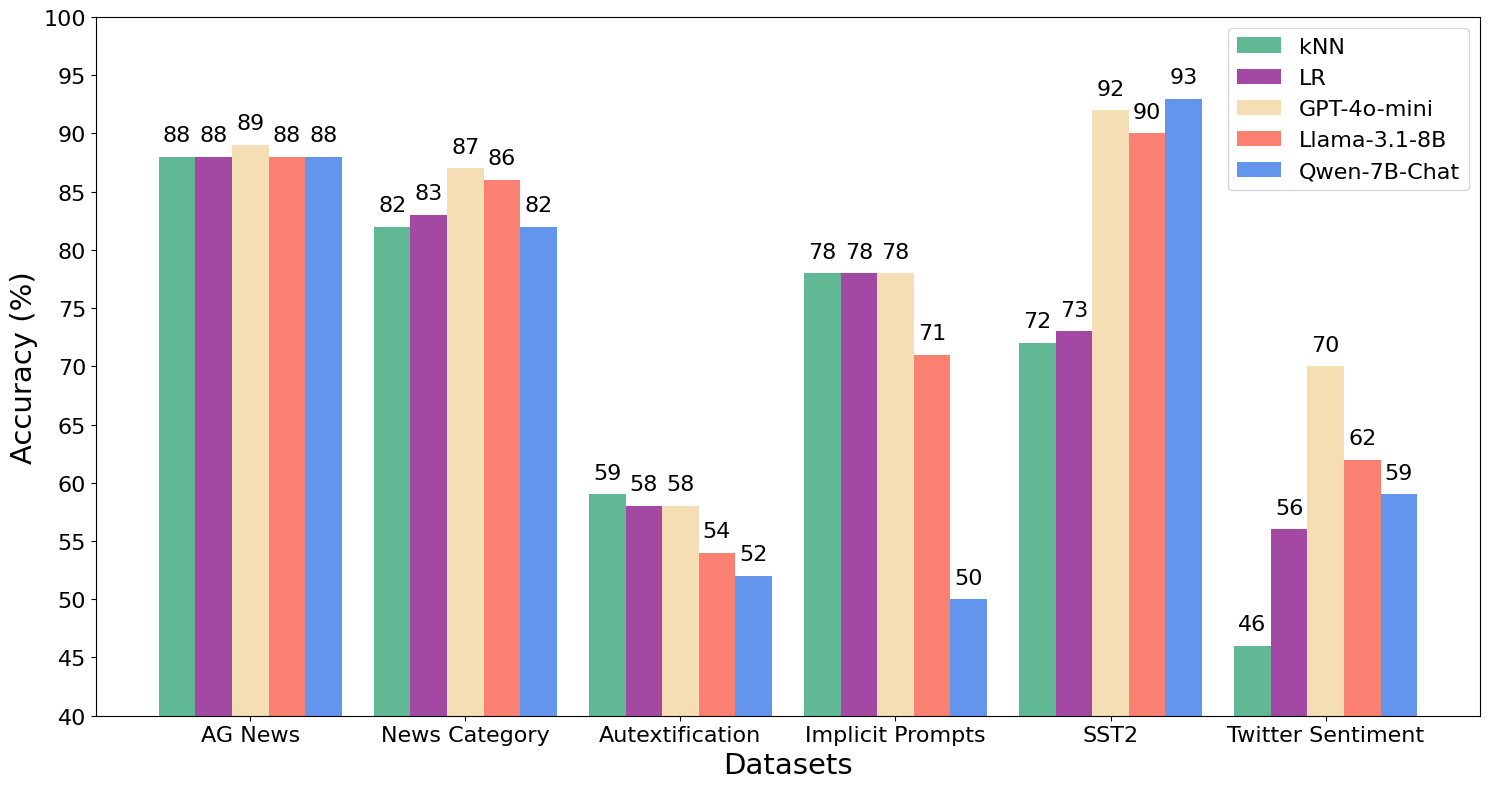}
  \caption{Performance on six datasets of kNN, LR, GPT-4o-mini, Llama-3.1-8B-Instruct, and Qwen-7B-Chat models for the same $k=10$ examples used by each model. In several datasets, LLMs perform similarly to kNN and LR, but this is not always the case.}
  \label{fig:experiments}
\end{figure}

In this paper, we continue the quest to understand ICL's behavior in the context of text classification tasks. Unlike previous work, we base our work on the intuitive empirical observation that ICL {\em should} be similar to supervised classifiers trained on ICL examples.\footnote{For brevity, we refer to these classifiers trained on ICL examples simply as \textit{classifiers} from here on. We use {\em example} or {\em demonstration} interchangeably to point to the demonstrations used in ICL or the training examples used by the classifiers.}
We further refine this idea into two directions.
First, building on the above hypothesis that ICL may behave similar to GD, these classifiers could belong to the family of classifiers based on gradient descent. To explore this direction, we leverage Logistic Regression (LR) \cite{lavalley2008logistic}.
Second, we observe that the decoder's attention mechanism feels similar to the example weights assigned by weighted k-nearest-neighbors (kNN) \cite{5408784}. Based on this, we explore two variants of kNN classifiers. 
We convert this intuition into three research questions: \textit{(1) Do LLMs with ICL behave similarly to classifiers trained on the same examples? (2) If so, which classifiers are closer, those based on gradient descent or those based on kNN? (3) When they do not behave similarly, what conditions make them behave differently?}

To answer the above questions, we compare the performance and behavior of three LLMs with ICL (GPT-4o-mini \cite{openai2024gpt4ocard}, Llama-3.1-8B-Instruct \cite{grattafiori2024llama3herdmodels}, and Qwen-7B-Chat {\cite{yang2024qwen2technicalreport}) against kNN and LR on six datasets. For a fair comparison, all models have access to the same training examples/demonstrations for each prediction. These examples were selected based on cosine similarity with the corresponding test example, using contextualized embeddings generated by Sentence-BERT \cite{reimers-2019-sentence-bert}. In this paper, we focus on text-classification tasks due to their ubiquity and ease of manual analysis.

We draw several observations from our analysis. First, in most situations, LLMs obtain similar or close performance to the classifiers (see the three left-most datasets in Figure~\ref{fig:experiments}). More interestingly, LLMs {\em behave} similarly to kNN and LR in these situations, as supported by a high Kappa agreement \cite{cohen1960coefficient} between their outputs. 

Second, LLMs show a slightly higher Kappa agreement with kNN, giving more empirical weight to the hypothesis that ICL behaves more similarly to the kNN mechanism.\footnote{However, please refer to Limitations for an additional discussion.}

Third, these observations do not always hold (see the three right-most datasets in Figure~\ref{fig:experiments}). 
Our hypothesis is that the similarity in behavior is driven by the relevance of ICL examples, where {\em example relevance} characterizes the extent to which ICL examples align with the testing data point.\footnote{For simplicity, henceforth we refer to example relevance as `relevance'.} To verify the hypothesis, we propose the following experiment: (a) We first annotate examples based on their relevance to the corresponding data point. The annotation is performed both manually and automatically. (b) Using this relevance, we determine whether its values are associated with the similarity between classifiers and LLMs. To do this, we compute the Kappa agreement between the classifier and LLM outputs. (c) We then examine the correlation between the Kappa values and the relevance scores. As anticipated, we observe a high correlation between relevance (both manual and machine) and Kappa agreement between models. In other words, when the relevance score is high, it naturally aligns with a higher Kappa agreement. This suggests that LLMs with ICL behave similarly to classifiers when the underlying examples are of high relevance. When not, their performance and behavior begin to diverge.

Beyond the more theoretical understanding of the behavior of ICL, our analysis yields an important practical insight: when relevant examples are available for a given task, LLMs can be replaced by a simpler and much more efficient classifier.

\section{Related Work}
\label{related-work}

We organize our discussion of previous work into two categories: papers that aim to understand how ICL works and those that improve its performance. 

\subsection{Understanding In-context Learning}
Several research studies have focused on ways to understand ICL in different scenarios. In the context of gradient descent (GD), \citet{dai2023gptlearnincontextlanguage} interpreted ICL as meta-optimizers through comparative analysis between ICL and explicit fine-tuning. \citet{akyürek2023learningalgorithmincontextlearning} theoretically proved that transformers can implement learning algorithms for linear models using gradient descent and closed-form ridge regression. 

Around the same time, \citet{pmlr-v202-von-oswald23a} explored the relation between training transformers on auto-regressive objectives and gradient-based meta-learning formulations. They investigated how transformers construct a loss function from given examples and the mechanisms through which they incorporate knowledge using the gradients of this loss function. Building on this idea, \citet{vonoswald2024uncoveringmesaoptimizationalgorithmstransformers} explored the underlying mechanisms of gradient-based mesa-optimization, i.e., a form of learning that ``is acquired through training, as opposed to being inherent to the model''
in autoregressive transformers trained for sequence modeling tasks. By reverse-engineering these models, they demonstrated that their in-context learning (ICL) capability arises from the mesa-layer, an innovative attention layer designed to efficiently solve a least-squares optimization problem. 
In the same space, \citet{zhou-etal-2025-context} analyzes whether LLMs perform ICL through an error-driven process similar to how humans learn. By testing for the inverse frequency effect, a phenomenon where rare examples more influence an agent than likely ones, the authors show that LLMs exhibit similar behavior to humans, suggesting that ICL involves an implicit error-driven learning mechanism like gradient descent.

In contrast, \citet{deutch2024incontextlearninggradientdescent} reexamined the hypothesis that GD approximates ICL, identifying key issues in the evaluation metrics and baselines proposed by \citet{dai2023gptlearnincontextlanguage}. \citet{shen2024pretrainedtransformerslearnincontext} similarly points out the inconsistent behavior of ICL and GD in a natural setting. They demonstrated that the hand-constructed weights used in these studies show properties that differ from those found in practical scenarios. \citet{fu2024transformerslearnachievesecondorder} provided evidence that transformers acquire in-context learning (ICL) capabilities by employing higher-order optimization instead of gradient descent (GD). They demonstrated that transformer architectures can efficiently implement Newton’s method, in which transformers significantly outperformed GD in terms of speed, exhibiting exponential acceleration.
\citet{nafar-etal-2025-learning} investigates the underlying mechanisms of ICL in LLMs within regression tasks. The work explores how LLMs learn from examples given in prompts versus relying on their internal knowledge.

Similar to most of these efforts, our work focuses on understanding ICL. We empirically observe that ICL behaves similarly to both gradient descent and kNN classifiers trained on the same examples as the demonstrations provided in ICL prompts. However, this observation only holds when the relevance of demonstrations is high. 
We believe this is more intuitive and has more practical implications.

\subsection{Improving In-context Learning}
Another line of research, such as \citet{wang2024largelanguagemodelslatent}, examines the in-context learning phenomenon from a Bayesian perspective, viewing real-world LLMs as latent variable models. Based on this concept, the work proposes a practical and better demonstration selection algorithm using a small LLM to approximate optimal latent concepts, which are then generalized to larger LLMs. The method yields significant performance gains, bridging the gap between theory and practical LLM behavior.
Focusing on the quality of examples, \citet{an2023skillbasedfewshotselectionincontext} proposes SKILL-KNN, which is a training-free, skill-based approach for few-shot example selection in in-context learning. The work utilizes prompting LLMs to generate task-relevant skill descriptions from raw inputs. SKILL-KNN mitigates biases associated with surface-level features commonly found in embedding-based selection methods to generate better examples. Whereas this prior research primarily uses kNN to find better examples for ICL, our work treats kNN as a comparative framework to gain insights into understanding the behavior of ICL.

Another work proposes framing the selection of demonstration examples as a sequential decision-making problem and employs reinforcement learning (RL) to learn policies for selecting examples \cite{zhang2022activeexampleselectionincontext}. This method is runtime expensive and does not consider examples based on similarity to the test data point. Instead, they observe that label distribution in the demonstration pool matters. In contrast, our work is simple, i.e., we use a standard model, Sentence-BERT, to provide a range of quality examples based on similarity. Further and more importantly, our work focuses on how the relevance of the examples impacts the decision of the downstream models. 

\citet{min2022rethinkingroledemonstrationsmakes} focuses on the structure of demonstrations rather than demonstration relevance for the test data. This work shows that factors like formatting and input-label alignment significantly affect LLM performance.
Nevertheless, this approach is different than our work, as our main focus is to investigate the importance of demonstration relevance.

Additionally, \citet{margatina2023activelearningprinciplesincontext} shows that selecting the semantically similar examples for prompts—using active learning strategies—can enhance ICL performance more than random demonstrations. It shows that the semantically similar demonstrations outperform other strategies and provide high-quality examples. Our paper takes this idea a step further. We aim to understand what happens when we use Sentence-BERT to select examples. Our work is different because we observe two important and practical conclusions when we use such examples: (i) ICL behaves very closely to kNN, which is novel, in cases of relevant examples; (ii) if the relevance of examples is not high, we do not observe the above correlation. This suggests that LLMs may back off to their existing knowledge in this situation.

\section{Empirical Analysis}
\label{sec:empirical}

Despite the progress in understanding the behavior of ICL, a critical gap remains in how these systems behave the way they do. We design our goal of understanding ICL's behavior as a comparison to classifiers trained on the same examples as ICL demonstrations. 
We approach this by performing a formal empirical analysis of the two, aimed at analyzing their similarity in performance and outputs across multiple models and datasets. Our findings revealed that ICL and classifiers are similar in performance, and in most cases, they generate the same labels.

The observation sparked our curiosity:  When they do not behave similarly, what conditions are correlated with differences in behavior? To explore this, we annotate a sample of ICL examples with respect to their relevance and compute the correlation between example relevance and behavior similarity. The results are striking; we observe that LLMs behave similarly to the classifiers when the relevance of the demonstrations is high. We further investigate how changes in the relevance of examples influence the similarity in behavior.

\subsection{Preliminaries}

This section outlines the experimental setup used to support our findings. To provide a comprehensive overview, we perform text classification tasks on six datasets with kNN, LR, and three LLMs (GPT, Llama, and Qwen). Prior to running the classification experiments, we generate text embeddings for both the training and test inputs of each dataset using Sentence-BERT \citep{reimers-2019-sentence-bert}.

\subsubsection{Models}

\paragraph{Sentence-BERT: } 

The model generates embeddings that are used to retrieve the closest $k$ examples from the training set based on similarity scores for each test case. To achieve this, we generate and save the contextualized embeddings for both the training and test data of each dataset using Sentence-BERT. 
During evaluation, we select the top-$k$ training examples with the highest similarity by sorting them in descending order of cosine similarity between their embedding vectors and the corresponding test embedding.
Note that, to have a fair comparison between classifiers and LLMs with ICL, the same examples were used by all LLMs.\footnote{Our goal in this paper is not to develop a novel example selection strategy, but rather to understand the classifiers and LLM behavior under a shared, common strategy. It is quite possible that both classifiers and LLM absolute performance would improve with a more sophisticated example selection strategy.}

\paragraph{kNN:} 
\label{knn}
We perform our experiment with two variants of kNN: \textit{unweighted kNN} \citep{fix1951discriminatory} and \textit{weighted kNN} \citep{5408784}. In the unweighted case, the predicted label for a test point is determined by a simple majority vote, meaning the label most common among its nearest neighbors is chosen. In weighted kNN, however, each neighbor’s vote is scaled by its similarity score, giving closer or more similar examples greater influence in the decision. In our experiments, we observed that both approaches behave quite similarly, with nearly identical performance outcomes. Hence, for simplicity, we report the results of correlation using only the unweighted kNN.\footnote{The performance results of weighted kNN are provided in Appendix~\ref{sec:weightedknn}.}

\paragraph{LR:} Similar to kNN, we train a \textit{Logistic Regression} model using labeled training data with ICL demonstrations when $k >  1$. In the case of a single training example—a situation that this discriminative classifier cannot handle—the model is configured to behave as a 1-NN classifier to keep the experiments consistent with kNN. We consider LR, as it is inspired by prior work on probing models that employ lightweight classifiers \cite{hewitt2019designinginterpretingprobescontrol, zhao-bethard-2020-berts}.

\paragraph{LLMs:} We used three pre-trained LLMs: GPT-4o-mini \cite{openai2024gpt4ocard}, Llama-3.1-8B-Instruct \cite{grattafiori2024llama3herdmodels}, and Qwen-7B-Chat \cite{yang2024qwen2technicalreport}. These LLMs have demonstrated proficiency across a range of tasks, such as text classification and sentiment analysis, using only a limited set of task-specific examples. We perform text classification using a common prompt with small changes based on the behavior of the LLM (a prompt example is provided in Appendix~\ref{sec:appendixC}).
The LLMs range in size from 7B to 8B. As mentioned, all LLMs used exactly the same ICL demonstrations as the other classifiers. 

\subsubsection{Datasets}

For our experiments, we used six datasets: AG News \citep{NIPS2015_250cf8b5}, News Category \citep{misra2022news}, AuTexTification \citep{sarvazyan2023overviewautextificationiberlef2023}, Implicit Prompts \citep{10786820}, SST-2 \citep{socher-etal-2013-recursive}, and SemEval-2019 Task 6 \citep{zampieri-etal-2019-semeval}. During evaluation, for datasets with large test partitions, we conduct experiments using a sample of approximately 1,000 test data points. Table~\ref{tab:accents} in Appendix~\ref{sec:appendixB} shows the total number of test cases along with the selected sample size and the number of labels. Additionally, the labels for these datasets are listed in Appendix~\ref{sec:appendixB} in Table~\ref{tab:labels}.

\subsection{Comparing Model Behavior}
\label{compare-models}

We begin with the analysis that motivates our study, highlighting the first research question: \textbf{RQ1: Do LLMs with ICL behave similarly to classifiers trained on the same examples?}
To explore this, we set out to compare the behavior of three different LLMs (GPT-4o-mini, Llama-3.1-8B-Instruct, and Qwen-7B-Chat) with classifiers (kNN and LR).

\subsubsection{Measuring Performance in Terms of Accuracy}
We conduct a systematic analysis of all the models to measure performance, which is evaluated on different datasets with the same $k$ examples. On examining their performance from Table~\ref{accuracy}, we see that the accuracy of kNN, LR, and three LLMs is similar in most cases. This is further visualized in a heatmap plot (Figure~\ref{fig:knn_accuracy_compare}) that reflects the performance differences between kNN and each LLM, as well as between LR and each LLM, across all settings. The figure indicates that the accuracy differences between classifiers and the three LLMs are minuscule on three datasets (left part of the figure). For example, the performance difference on AG News and News Category is $0$ or very close to $0$ in most cases.
However, these differences increase significantly on the three datasets in the right part of the figure, with the difference reaching as high as $35\%$ between kNN and Qwen on Implicit Prompts. 

\setlength{\tabcolsep}{1.3pt} 
\renewcommand{\arraystretch}{1.7} 
\begin{table*}[!t]
  \centering
  \fontsize{8pt}{8pt}\selectfont
  \begin{tabular}{l>{\columncolor{blue!5}}c
  >{\columncolor{blue!5}}c
  >{\columncolor{blue!5}}c
  >{\columncolor{blue!5}}c
  >{\columncolor{blue!5}}c
  >{\columncolor{yellow!10}}c
  >{\columncolor{yellow!10}}c
  >{\columncolor{yellow!10}}c
  >{\columncolor{yellow!10}}c
  >{\columncolor{yellow!10}}c
  >{\columncolor{red!8}}c
  >{\columncolor{red!8}}c
  >{\columncolor{red!8}}c
  >{\columncolor{red!8}}c  
  >{\columncolor{red!8}}c  
  >{\columncolor{green!8}}c
  >{\columncolor{green!8}}c
  >{\columncolor{green!8}}c
  >{\columncolor{green!8}}c
  >{\columncolor{green!8}}c}
    \toprule
    \multirow{2}{*}{\textbf{Dataset}} 
      & \multicolumn{5}{>{\columncolor{blue!5}}c}{$k=1$} 
      & \multicolumn{5}{>{\columncolor{yellow!10}}c}{$k=10$} 
      & \multicolumn{5}{>{\columncolor{red!8}}c}{$k=20$} 
      & \multicolumn{5}{>{\columncolor{green!8}}c}{$k=30$} \\
    \cmidrule(lr){2-6}
    \cmidrule(lr){7-11}
    \cmidrule(lr){12-16}
    \cmidrule(lr){17-21}
      & kNN & LR & GPT & Llama & Qwen & kNN & LR  & GPT & Llama & Qwen & kNN & LR  & GPT & Llama & Qwen & kNN & LR  & GPT & Llama & Qwen \\
    \midrule
    AG News & 0.85 & 0.85 & 0.83 & 0.84 & 0.85 & 0.88 & 0.88 & 0.89 & 0.88 & 0.88 & 0.88 & 0.89 & 0.89 & 0.88 & 0.89 & 0.87 & 0.89 & 0.89 & 0.89 & 0.88\\
    News Category & 0.80 & 0.80 & 0.86 & 0.81 & 0.73 & 0.82 & 0.83 & 0.87 & 0.86 & 0.82 & 0.82 & 0.84 & 0.87 & 0.87 & 0.82 & 0.82 & 0.83 & 0.88 & 0.88 & 0.82 \\
    AuTexTification & 0.54 & 0.54 & 0.52 & 0.55 & 0.54 & 0.59 & 0.58 & 0.58 & 0.54 & 0.52 & 0.58 & 0.58 & 0.58 & 0.55 & 0.48 & 0.59 & 0.58 & 0.55 & 0.54 & 0.52\\
    Implicit Prompts & 0.69 & 0.69 & 0.74 & 0.58 & 0.41 & 0.78 & 0.78 & 0.78 & 0.71 & 0.50 & 0.79 & 0.78 & 0.78 & 0.74 & 0.44 & 0.79 & 0.79 & 0.78 & 0.73 & 0.48\\
     SST-2 & 0.68 & 0.68 & 0.92 & 0.80 & 0.94 & 0.72 & 0.73 & 0.92 & 0.90 & 0.93 & 0.73 & 0.73 & 0.92 & 0.92 & 0.93 & 0.73 & 0.74 & 0.92 & 0.92 & 0.94\\  
     SemEval-2019 & 0.39 & 0.39 & 0.58 & 0.51 & 0.49 & 0.46 & 0.56 & 0.70 & 0.62 & 0.59 & 0.48 & 0.61 & 0.73 & 0.65 & 0.57 & 0.47 & 0.63 & 0.74 & 0.66 & 0.56\\
    \bottomrule
  \end{tabular}
  \captionof{table}{\label{accuracy}
    Test accuracy on six datasets for unweighted-kNN, LR, and three LLMs, using $1$, $10$, $20$, and $30$ examples.
  }
\end{table*}

\begin{figure}[!h]
 \centering
  \includegraphics[width=\columnwidth]{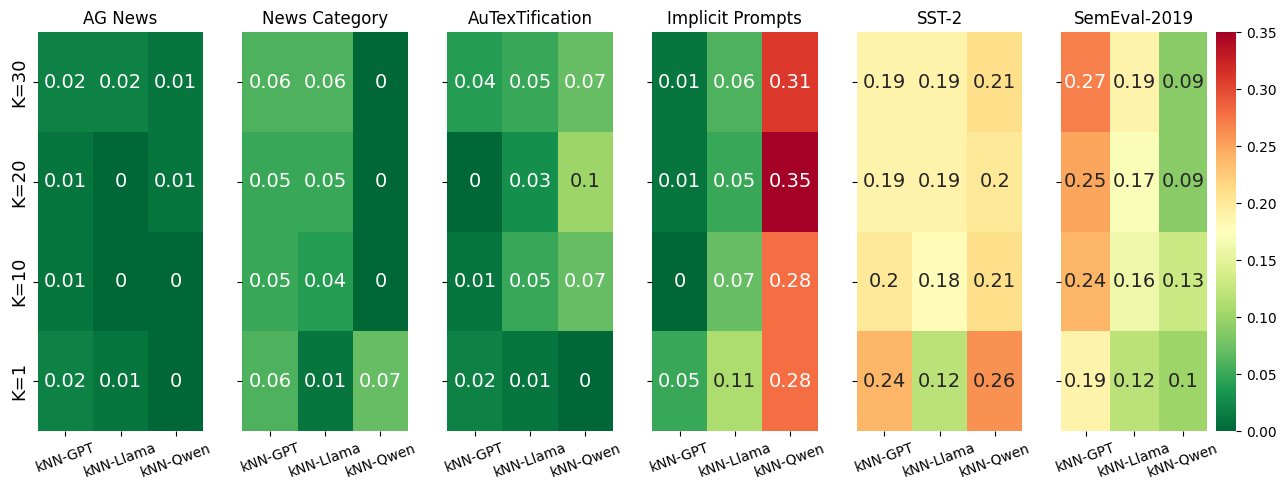}
  \par\smallskip
  \includegraphics[width=\columnwidth]{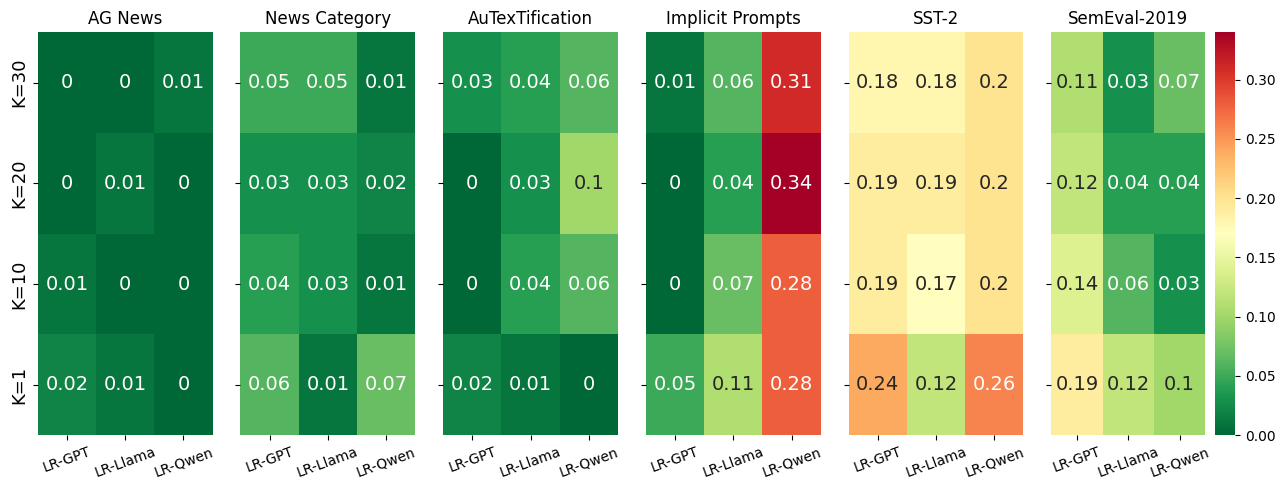}
  \caption{Visualization highlighting the difference in accuracy between kNN (top) and LR (bottom), each compared with three LLMs. Each vertical block is a different dataset; each row is a setting with a different number of ICL examples (from $1$ to $30$). The difference is also encoded with a color gradient, where green means lower difference (higher similarity) and red indicates high differences (low similarity).
  }
  \label{fig:knn_accuracy_compare}
\end{figure}

Building on these findings, a natural next question arose;  \textbf{RQ2: If ICL behaves similarly to classifiers with the same set of examples, which classifiers are closer, those based on gradient descent (GD) or those based on k-nearest neighbors (kNN)?} We observe that kNN and Logistic Regression have comparable performance, while ICL aligns more closely with kNN.\footnote{Detailed comparison of kNN with LR is given in Appendix~\ref{sec:knnvslr}.} However, as we have a very simple gradient-based classifier, the results may vary with non-linear GD classifiers. 

\subsubsection{Assessing Model Similarity Using Contingency Matrices}
\label{sec:contingency}

At this point, one might reasonably argue that performance parity among models does not necessarily imply similar behavior, as they may achieve comparable performance scores while exhibiting different predictions of labels. To better understand the similarity in performance, we next construct contingency matrices between each LLM and the classifiers' outputs. This helps to investigate whether classifiers and LLMs actually {\em behave} in a similar manner, i.e., \textit{do classifiers and LLMs generate identical labels given the same input?}

In our work, we plot the contingency matrices between kNN and LR (X-axis) and the three LLMs (Y-axis) on the six datasets. These matrices provide a structured way to compare and evaluate the labels predicted by different models. That is, the more similar two outputs are, the more weight the diagonal will have. 
The matrices support our observation: classifiers and LLMs exhibit similar behavior on some datasets but not on all.
For example, referring to the contingency matrices of AG News (Figure~\ref{fig:agnews}) and News Category (Figure~\ref{fig:newscat}), we find that they follow diagonal dominance, which indicates that both classifiers produce almost the same labels as LLMs. In an opposing case of Implicit Prompts (Figure~\ref{fig:implicit}), we observe that the classifiers tend to favor the offensive label. We hypothesize that this occurs because the label distribution of the dataset is heavily biased towards offensive prompts.\footnote{All contingency matrices and extended discussion included in Appendix~\ref{sec:addcontingency}.}

\begin{figure}[!h]
 \centering
        \includegraphics[width=0.19\textwidth]{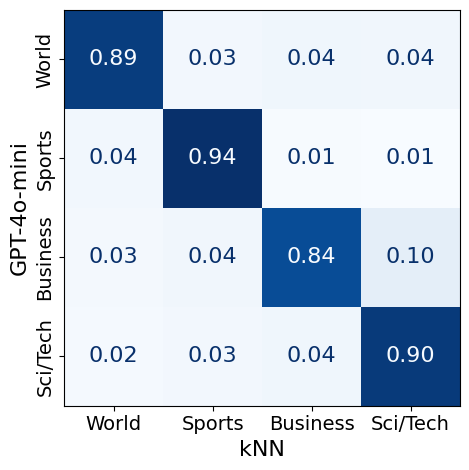}
        \includegraphics[width=0.19\textwidth]{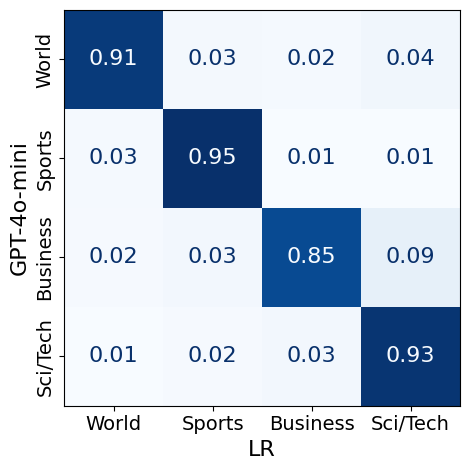}
        \par\smallskip
        \includegraphics[width=0.19\textwidth]{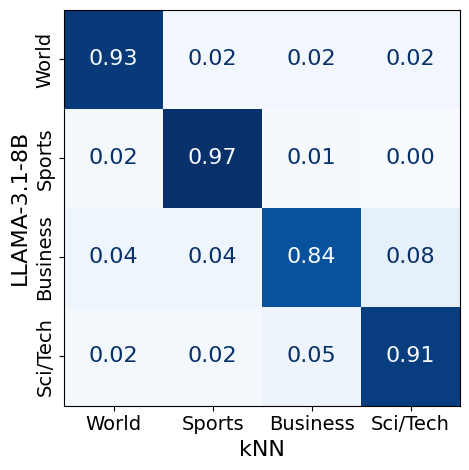}
        \includegraphics[width=0.19\textwidth]{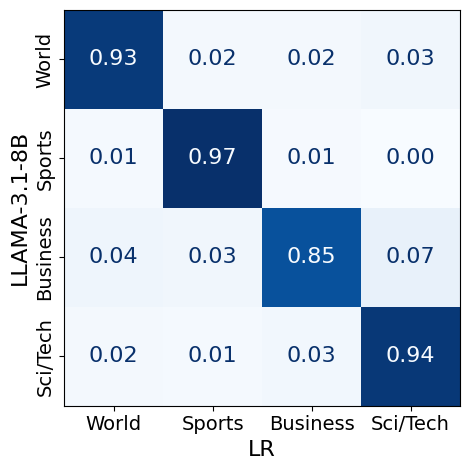}
        \par\smallskip
        \includegraphics[width=0.19\textwidth]{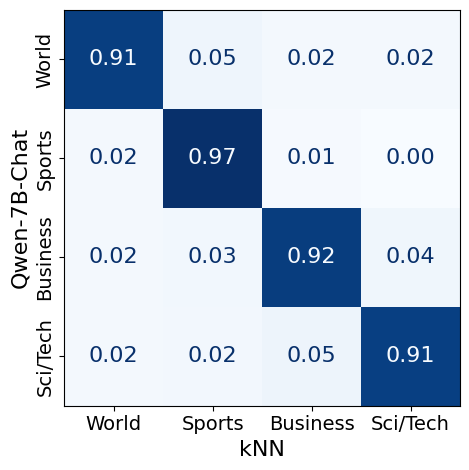}
        \includegraphics[width=0.19\textwidth]{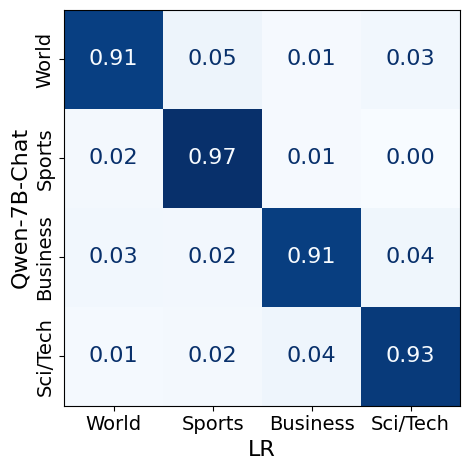}
        \caption{Contingency matrices for AG News: Comparison of kNN vs GPT  (top-left), kNN vs Llama (middle-left), and kNN vs Qwen (bottom-left) and LR vs GPT  (top-right), LR vs Llama (middle-right), and LR vs Qwen (bottom-right) for $k=20$.}
  \label{fig:agnews}
\end{figure}

\begin{figure}[!h]
 \centering
        \includegraphics[width=0.23\textwidth]{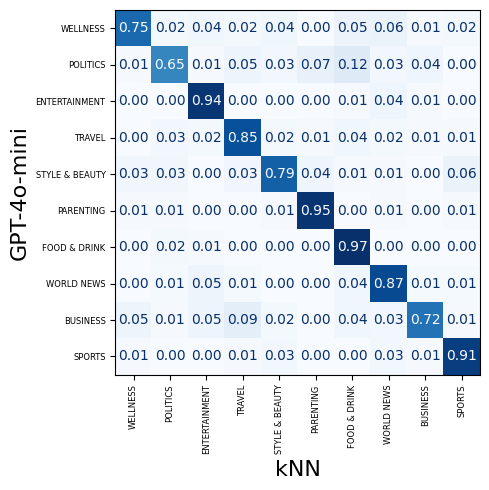}
        \includegraphics[width=0.23\textwidth]{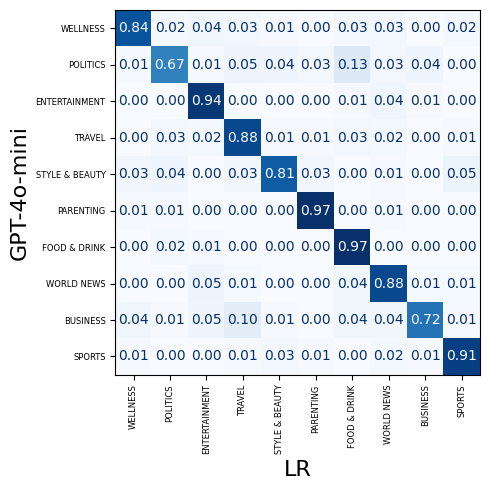}
        \par\smallskip
        \includegraphics[width=0.23\textwidth]{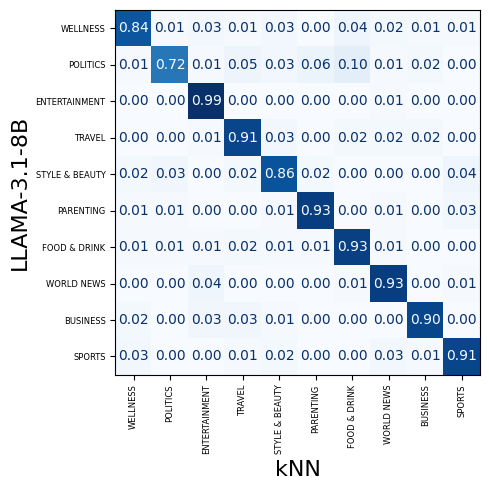}
        \includegraphics[width=0.23\textwidth]{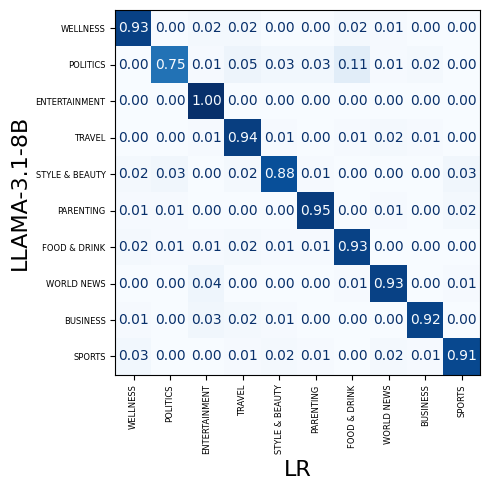}
        \par\smallskip
        \includegraphics[width=0.23\textwidth]{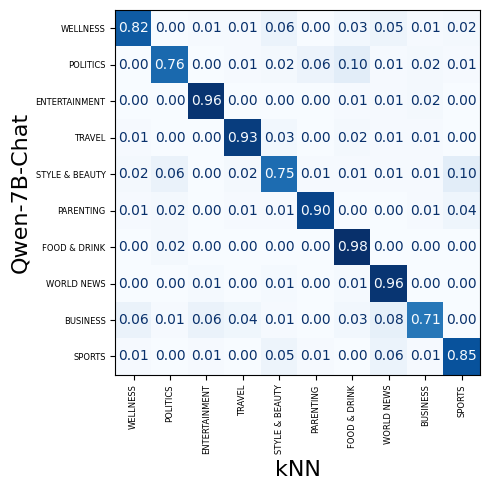}
        \includegraphics[width=0.23\textwidth]{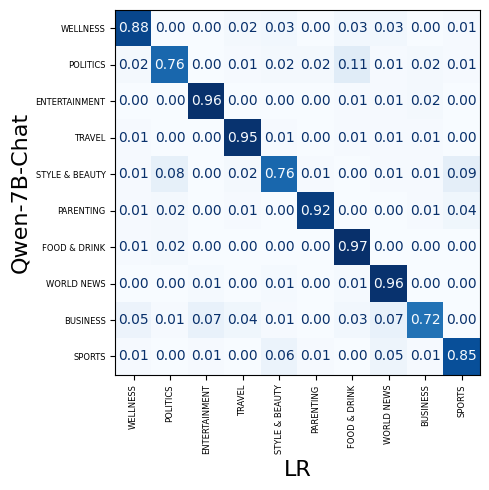}
        \caption{Contingency matrices for News Category: The same configuration follows from Figure~\ref{fig:agnews}.}
  \label{fig:newscat}
\end{figure}

Overall, we observe that diagonal dominance is excellent in $2$ out of $6$ datasets, good in $2$, and decent in $2$.

\subsection{Does Example Relevance Correlate with Behavior Similarity?}
\label{correlation-analysis}

All the previous observations immediately indicate a puzzling conclusion: classifiers and LLMs with ICL obtain similar performance in general but not in all cases. 
These results lead to a new question: \textbf{RQ3: When LLMs and classifiers do not behave similarly, what conditions are correlated with differences in behavior?}
To investigate this, we hypothesize that, given the performance of kNN is driven by examples, it is essential to analyze the correlation between example relevance and the performance of models.

\subsubsection{Understanding Example Relevance}

In our context, we define {\em example relevance} in terms of {\em task relevance for the provided evaluation data point}. Informally, relevance indicates how well a semantically similar example closely matches the input test data prompt for the task at hand. By selecting examples that are relevant to the input prompt, we can measure whether a direct correlation between example relevance and the similarity in behavior of models exists. 

To establish correlation, we mark the top-$k$ semantically similar examples with a Boolean label for each example. We assign a value of `$1$' to the examples that are relevant to the input test data prompt, and `$0$' to those that are not. For each test data prompt, we determine an average of these values for all examples, which we call a {\em relevance score}, reflecting the percentage of the top-$k$ examples that are relevant to the input prompt.

For labeling the examples, we consider two options: one based on manual annotations and the other based on the respective LLM. In the first case, considering human annotators for all examples across the entire test prompts is both challenging and expensive. For this reason, we sample and annotate a set of 50 test data prompts for each setting of $k=1, 10$, and $20$ examples manually by Human-A.\footnote{We annotated a total of $50 \times (20 + 10 + 1) = 1,550$ examples per dataset.} To verify the robustness of the manual annotations, a smaller subsample of data points was independently annotated by another human annotator (Human-B).\footnote{The two annotators each annotated two data points per configuration per dataset, for a total of $2 \times (20 + 10) = 60$ examples per dataset.}
We find substantial Kappa agreement ($\approx 70\%$) of relevance scores between the two human annotators. 

We use the same set of $1,550$ examples, this time using the respective LLM to calculate their corresponding relevance score. We then compare the agreement between relevance scores by human and machine annotators.
To measure agreement, we introduce two scores. The first score, \Krel\  represents the Kappa agreement on relevance between a human and an LLM. The second measures the inclusion score (\Inc), which reflects the fraction of humans’ relevance scores that is included within that of LLMs. \Inc\ score of $0$ indicates no inclusion, and a score of $1$ indicates complete inclusion.
Table~\ref{kappa-relevance} shows the Kappa (\Krel), and Inclusion  (\Inc) scores of relevance between Human-A and Gpt-4o-mini.

\setlength{\tabcolsep}{1.2pt}
\renewcommand{\arraystretch}{1.1}
\begin{table}[!h]
\fontsize{9pt}{9pt}\selectfont
\centering
  \begin{tabular}{p{2.2cm}cccccc}
    \toprule
    \multirow{2}{*}{Datasets}    & \multicolumn{2}{c}{$k=1$} & \multicolumn{2}{c}{$k=10$} & \multicolumn{2}{c}{$k=20$}\\
    \cmidrule(lr){2-3}
    \cmidrule(lr){4-5}
    \cmidrule(lr){6-7}
    &  \Krel & \Inc  &  \Krel & \Inc &  \Krel & \Inc  \\
    \midrule
    AG News & 0 & 1 & 0.05 & 0.98 & 0.09 &  0.97\\
    News Category & 0.41 & 0.95 & 0.28 & 0.92 & 0.24 & 0.89 \\
    AuTexTification & 0.52 & 0.92 & 0.25 & 0.76 & 0.23 & 0.73\\
    Implicit Prompts & -0.03 & 0.98 & 0.04 & 0.98 & 0.04 & 0.97 \\
    SST-2 & 0.09 & 1 & 0.06 & 0.99 & 0.05 & 0.98 \\
    SemEval-2019 & 0.38 & 0.8 & 0.26 & 0.91 & 0.17 & 0.89   \\
    \bottomrule
  \end{tabular}
  \caption{\label{kappa-relevance}
    Kappa agreement between relevance of Human-A and Gpt-4o-mini (\Krel) and corresponding Inclusion score (\Inc) for increasing $k$.
  }
\end{table}

We find that even though the \Krel\ value is not good, the \Inc\ reflects an interesting conclusion that a vast majority of examples annotated as relevant by the human annotator were also considered relevant by the LLM annotator. 
In other words, the human annotators are more conservative than the LLMs in labeling a demonstration as relevant.
To compensate for this, we use two types of relevance scores (Human and LLM) in the following sections for our correlation analyses.
For the LLM annotator, which scales better, we also generate the relevance scores for {\em all} test data prompts for $k=1, 10$, and $20$ examples.

\subsubsection{Correlation between Example Relevance and Model Similarity}
\label{sec:correlation}

Based on prior discussions, we know that ICL is similar to classifiers in many situations; the next step is to identify the conditions under which this similarity occurs. To explore this, we examine how the relevance scores of humans and LLMs are correlated with the Kappa agreement between models. 
For simplicity, we refer to the Kappa agreement between the labels predicted by the classifiers and LLM as model Kappa, denoted by \Kmodel.

To measure the correlation between model similarity and example relevance, we employ two types of correlations that are widely used to measure the association between two variables: the Pearson correlation coefficient ($r$) and the R-squared ($R^{2}$) coefficient \citep{sedgwick2012pearson, wright1921correlation}. $r$ captures the linear correlation between two variables by computing the ratio between the covariance of two variables and the product of their standard deviations; it ranges between $-1$ and $1$. $R^{2}$ is the proportion of the variation in one variable that is predictable from the other one. For example, we interpret the $R^{2}$ between example similarity and the Kappa agreement between two models as the fraction of the Kappa score that can be explained by example relevance.

Formally defining, we measure the correlation between the \Kmodel\ and two variables: the first one is the average relevance score determined by a human (denoted by \Rhum), and the second is by the respective LLM as an annotator (denoted by \Rllm). Intuitively, we hypothesize that there should be a direct correlation between both relevance scores and model Kappa. In other words, higher example relevance should correspond to a higher Kappa agreement score.\footnote{Since most of the datasets involve multi-class classification, we use the multi-class Kappa agreement formula \citep{cohen1960coefficient}.}

\setlength{\tabcolsep}{4pt}
\renewcommand{\arraystretch}{1.1}
\begin{table*}[!t]
\fontsize{10pt}{10pt}\selectfont
\centering
  \begin{tabular}{p{3.2cm}ccccccccc}
    \toprule
    \multirow{2}{*}{Datasets \textbackslash{}  Neighbors}    & \multicolumn{3}{c}{$k=1$} & \multicolumn{3}{c}{$k=10$} & \multicolumn{3}{c}{$k=20$}\\
    \cmidrule(lr){2-4}
    \cmidrule(lr){5-7}
    \cmidrule(lr){8-10}
    & \Rhum & \Kmodel & \Rllm  & \Rhum & \Kmodel & \Rllm & \Rhum & \Kmodel & \Rllm\\
    \midrule
     AG News  & 0.9 & 0.87 & 1 & 0.794 & 0.87 & 0.976 & 0.7 & 0.87 & 0.952\\
     News Category & 0.8 & 0.87 & 0.88 & 0.664 & 0.87 & 0.834 & 0.553 & 0.84 & 0.795\\
     AuTexTification  & 0.5 & 0.06 & 0.66 & 0.312 & 0.03 & 0.546 & 0.267 & 0.02 & 0.507  \\
     Implicit Prompts & 0.96 & 0.19 & 0.98 & 0.756 & 0.39 & 0.966 & 0.627 & 0.54 & 0.956  \\
     SST-2  & 0.7 & 0.31 & 0.98 & 0.596 & 0.49 & 0.97 & 0.471 & 0.57 & 0.961  \\
     SemEval-2019  & 0.2 & 0.12 & 0.38 & 0.21 & 0.3 & 0.57 & 0.154 & 0.34 & 0.581  \\
    \midrule
    \rowcolor{lightgreen!45} $Correlation$-$A$ &  
    \multicolumn{3}{c}{\parbox[c]{3.2cm}{\centering $r$: 0.559 \\ [0.7mm] $R^{2}$: 0.313}} & 
    \multicolumn{3}{c}{\parbox[c]{3.2cm}{\centering $r$: 0.728 \\ [0.7mm] $R^{2}$: 0.530}} &  
    \multicolumn{3}{c}{\parbox[c]{3.2cm}{\centering $r$: 0.795  \\ [0.7mm] $R^{2}$: 0.631}} \\
    \midrule
    \rowcolor{lightblue!45} $Correlation$-$B$ &  
    \multicolumn{3}{c}{\parbox[c]{3.2cm}{\centering $r$: 0.528  \\ [0.7mm] $R^{2}$: 0.279}} & 
    \multicolumn{3}{c}{\parbox[c]{3.2cm}{\centering $r$: 0.668  \\ [0.7mm] $R^{2}$: 0.446}} &  
    \multicolumn{3}{c}{\parbox[c]{3.2cm}{\centering $r$: 0.768  \\ [0.7mm] $R^{2}$: 0.589}} \\
    \bottomrule
  \end{tabular}
  \caption{\label{GPT-50}
    Correlation comparison with \textbf{GPT-4o-mini} and \textbf{kNN} on $50$ data prompts.
    The top part of the table shows three columns: the human-annotated relevance score for ICL examples (\Rhum); Kappa agreement between the labels of classifier and LLM (\Kmodel); and the machine-annotated relevance score on ICL examples used (\Rllm). The bottom part shows two correlations; $Correlation$-$A$ = $Corr$(\Kmodel, \Rhum) (in green) and $Correlation$-$B$ = $Corr$(\Kmodel, \Rllm) (in purple).
  }
\end{table*}

\renewcommand{\arraystretch}{1.1}
\begin{table*}[!ht]
\fontsize{10pt}{10pt}\selectfont
\centering
  \begin{tabular}{p{3.5cm}cccccc}
    \toprule
    \multirow{2}{*}{Datasets \textbackslash{}  Neighbors}    & \multicolumn{2}{c}{$k=1$} & \multicolumn{2}{c}{$k=10$} & \multicolumn{2}{c}{$k=20$}\\
    \cmidrule(lr){2-3}
    \cmidrule(lr){4-5}
    \cmidrule(lr){6-7}
    & \Kmodel & \Rllm  & \Kmodel & \Rllm & \Kmodel & \Rllm\\
    \midrule
    AG News & 0.82 & 0.974 & 0.87 & 0.934 & 0.86 & 0.917 \\
    News Category & 0.67 & 0.924 & 0.67 & 0.839 & 0.69 & 0.792 \\
    AuTexTification & 0.16 & 0.608 & 0.09 & 0.511 & -0.02 & 0.468 \\
    Implicit Prompts & 0.3 & 0.868 & 0.33 & 0.866 & 0.28 & 0.861 \\
    SST-2 & 0.37 & 0.944 & 0.44 & 0.929 & 0.46 & 0.911 \\
    SemEval-2019 & 0.17 & 0.863 & 0.3 & 0.868 & 0.32 & 0.87 \\
    \midrule
    \rowcolor{red!10} $Correlation$-$B$ &  
    \multicolumn{2}{c}{\parbox[c]{3.2cm}{\centering $r$: 0.674  \\ [0.7mm]  $R^{2}$: 0.455}} & 
    \multicolumn{2}{c}{\parbox[c]{3.2cm}{\centering $r$: 0.676  \\ [0.7mm]  $R^{2}$: 0.457}} &  
    \multicolumn{2}{c}{\parbox[c]{3.2cm}{\centering $r$: 0.691  \\ [0.7mm]  $R^{2}$: 0.477}} \\
    \bottomrule
  \end{tabular}
  \caption{\label{GPT-test}
    Correlation comparison with \textbf{GPT-4o-mini} and \textbf{kNN} on $all$ data prompts.
    The table shows the correlation between: Kappa agreement between the outputs of classifier(kNN) and LLM (\Kmodel); and the annotated relevance score by LLM on ICL examples (\Rllm).
  }
\vspace{-2mm}
\end{table*}

Building on this hypothesis, we observe three key findings in Table~\ref{GPT-50}. First, the Pearson Correlation ($r$) values are positive for each correlation, confirming that the more relevant the examples are, the more similar the LLM and classifier behave. This holds for both types of correlations, $A$ and $B$, as defined in the table. 
Second, we find that increasing the number of examples ($k$) per test data point, the data more strongly supports these observations, suggesting that the LLM increasingly relies on ICL examples as more instances are provided in the context.
Third, focusing on the $R^{2}$ value for $Correlation$-$A$ for $k=20$, we indicate that up to $63.1\%$ of the behavior similarity can be explained by example relevance.\footnote{Tables~\ref{gpt-lr-50} through~\ref{qwen-lr-50} in Appendix~\ref{sec:appendixG} show the results of all the analyses of correlation for the three LLMs investigated with classifiers.}

\subsubsection{Scaling Up the Correlation Analysis to the Entire Test Partitions}
\label{llm-relevance}

As mentioned earlier, manual relevance is limited and expensive; hence, to scale up the experiment, we calculate the average relevance score on the annotations given by the underlying LLMs.
This implies that the same procedure is extended to all test prompts, with the three LLMs employed as annotators for the relevance score of examples.

Consistent with the observations in Table~\ref{GPT-50}, Table~\ref{GPT-test} shows that LLM maintains high relevance scores (\Rllm) across $k$ values because of high-quality semantically similar examples. 
Furthermore, the correlation value exhibits an upward trend, reflecting a similar pattern to that previously observed with both human and machine relevance on $50$ test prompts.
Note that in some cases, there is a slight drop in relevance scores from $k=10$ to $k=20$, which occurs because an increase in the number of examples introduces low-quality examples in the set, thereby reducing the overall relevance score. Because this drop aligns with model Kappa in these cases, the correlation values are not affected and exhibit the same trend as before.

\section{Discussion}

We hypothesize that when the relevance of the ICL examples is high, both LLMs and the classifiers rely on them for their predictions. When example relevance is low, differences begin to emerge; LLMs can back off to their parametric memory and still perform well in some cases. In contrast, the classifiers' performance is directly impacted by example relevance. Additionally, LLM indeed backs off to its parametric memory when using low-relevance demonstrations in ICL.\footnote{This was validated by comparing the outputs of the LLM with poor-quality examples with a zero-shot configuration of the same LLM. Please see Appendix~\ref{sec:poorvszero}.} 

We summarize the key observations from this study in Algorithm~\ref{alg:cap}. The algorithm indicates that when the examples selected to be used by the classifier or ICL in LLMs are relevant (which was established using both human and LLM decisions), ICL performs and behaves similarly to kNN (and a little less so to LR).
We validate this observation by showing that there is a high correlation between example relevance and similarity between ICL and the classifier. We also notice that when the relevance of examples is low, the two models diverge: the performance of classifiers generally drops, whereas the performance of the LLM is contingent on its pre-training and parametric memory, i.e., if the task at hand is well represented in its parametric memory, it continues to perform well. 

Similar to our analysis, the task recognition method of \citet{pan2023incontextlearninglearnsincontext} also makes the distinction between non-parametric and parametric memory. However, we explain why both are important, and how ICL may choose one or another depending on the relevance of the demonstrations.
``Task recognition'' encourages LLMs to back off to a certain part of the parametric memory, so we think that their observation complements nicely our own.

This summary suggests an important practical insight: in the situations where the relevance of the ICL examples is high, the LLM can be safely replaced by the simpler kNN algorithm, which can be efficiently implemented using vector databases such as FAISS \cite{douze2024faiss}. This, of course, puts the onus on the methods for the example selection, but this is research that has already shown progress \cite{akyürek2023learningalgorithmincontextlearning, zhou-etal-2025-context, shen2024pretrainedtransformerslearnincontext, fu2024transformerslearnachievesecondorder, nafar-etal-2025-learning, wang2024largelanguagemodelslatent, an2023skillbasedfewshotselectionincontext}.

\begin{algorithm}[!t]
\fontsize{10pt}{11pt}\selectfont
\caption{Summary of LLM with ICL behavior}
\label{alg:cap}
\begin{algorithmic}

\State $top_k \gets$ top $k$ semantically similar examples

\If{$top_k$ is relevant}
    \State LLMs $\approx$ kNN
\ElsIf{$top_k$ is less relevant}
    \State Performance of all classifiers drops
    \State Performance of LLMs may remain high, possibly\\~~~~~~~~ due to parametric memory information    
\EndIf
\end{algorithmic}

\end{algorithm}

\section{Conclusion}
In this paper, we examine how closely ICL behaves to classifiers based on kNN or GD trained on the same ICL examples. We find that ICL behaves more similarly to kNN when subjected to certain conditions driven by the relevance of examples. This suggests that if one can strategically select highly relevant examples, they can just use kNN. However, in cases where the examples are of lower relevance, LLMs demonstrate greater robustness to noise due to their ability to back off to their parametric memory. As a result, LLMs become a more suitable choice in such scenarios, as prior knowledge allows them to perform better.

\section{Acknowledgments}
Mihai Surdeanu declares a financial interest in lum.ai. This interest has been properly disclosed to the University of Arizona Institutional Review Committee and is managed in accordance with its conflict of interest policies.

\section{Limitations}
In our current work, we focus on three LLMs and six text classification datasets, which provide initial insights into the observed patterns. We believe our observations are supported empirically by our current work and are already actionable.
However, to strengthen our understanding of the underlying behavior of ICL, we plan to incorporate more LLMs, multiple datasets, and more NLP tasks in future work. This will help validate our conclusions with greater confidence, ensuring that our results hold across diverse scenarios.

Further, our comparison between kNN and gradient descent and ICL used a linear GD-based classifier (logistic regression). This is a common strategy in probing \cite{hewitt2019designinginterpretingprobescontrol, zhao-bethard-2020-berts}. However, it is possible that ICL behaves closer to a non-linear GD-based classifier, such as a feed-forward neural network.

One concern is whether there is a causal link between the two variables that are correlated. i.e., the relevance scores of demonstrations and the Kappa agreement between LLMs and classifiers. In future work, we propose to compare two identical configurations (i.e., same LLM, same dataset, same number of demonstrations $k$), with the only difference being that one has access to high-quality demonstrations while the other uses low-quality examples. If causality holds, the Kappa values would be high for all configurations, and they should drop when we use bad examples.

\section{Ethical Considerations}
This work involves some datasets that may contain offensive content, which could be distressing to some readers. We have taken measures by using appropriate disclaimers to acknowledge the offensive language.   

\bibliography{custom}

\appendix

\section{Analysis of Manual Annotation Quality}
\label{sec:appendixA}
Table~\ref{tab:annotator-agreement} presents the inter-annotator agreement for the manual annotations of example relevance. The Kappa value, representing the level of agreement between annotators, was computed based on the manual annotations of two data points per dataset for $k = 10$ and $k = 20$. In total, $2 \times (10+20) = 60$ individual examples per dataset were annotated by each annotator.

\setlength{\tabcolsep}{22pt}
\renewcommand{\arraystretch}{1}
\begin{table}[!ht]
  \begin{tabular}{p{30mm}p{15mm}}
    \toprule
    Nearest Neighbors & Kappa\\
    \midrule
    $k$ = 10 & 0.70\\
    $k$ = 20 & 0.68\\
    \bottomrule
  \end{tabular}
  \caption{Kappa agreement scores between the two annotators across all datasets.}
  \label{tab:annotator-agreement}
\end{table}

These values indicate substantial agreement between the two human annotators, which provides evidence that the manual annotations of example relevance 
are reliable.

\section{Dataset Labels \& Statistics}
\label{sec:appendixB}

The complete statistics of all datasets used in the experiments are presented in Table~\ref{tab:accents}.

\setlength{\tabcolsep}{6pt}
\renewcommand{\arraystretch}{1.2}
\begin{table}[!h]
\fontsize{9.5pt}{9.5pt}\selectfont
  \begin{tabular}{
  p{22mm}
  >{\centering\arraybackslash}p{8mm}
  >{\centering\arraybackslash}p{8mm}
  >{\centering\arraybackslash}p{10mm}
  >{\centering\arraybackslash}p{8mm}
}
    \toprule
    Datasets & Train & Total Test & Sampled Test & \#Labels \\
    \midrule
    AG News & 120000 & 7600 & 1001 & 4\\
    News Category & 40000 & 10000 & 1000 & 10\\
    AuTexTification & 33845 & 21832 & 1000 & 2\\
    Implicit Prompts & 6615 & 1655 & 1655 & 2\\
    SST-2  & 67349 & 872 & 872 & 2\\
    SemEval-2019 & 74682 & 1000 & 1000 & 4\\
    \bottomrule
  \end{tabular}
  \caption{Data statistics of all the datasets used.}
  \label{tab:accents}
\end{table}

Additionally, the labels for the six datasets used in this work are presented in Table~\ref{tab:labels}. We employ both binary and multi-class labels to ensure the generality of our approach.

\setlength{\tabcolsep}{8pt}
\renewcommand{\arraystretch}{1.5}
\begin{table}[!h]
\center
\fontsize{9.5pt}{9.5pt}\selectfont
  \begin{tabular}{p{22mm}p{42mm}}
    \toprule
    Datasets & Labels\\
    \midrule
    AG News & World, Sports, Business, Sci/Tech\\
    News Catgory & Wellness, Politics, Entertainment, Travel, Style \& Beauty, Parenting, Food \& Drink, World News, Business, Sports \\
    AuTexTification & Human, Generated\\
    Implicit Prompts & Offensive, Not Offensive\\
    SST-2 & Positive, Negative\\
    SemEval-2019 & Negative, Positive, Neutral, Irrelevant\\
    \bottomrule
  \end{tabular}
  \caption{Labels for the six datasets used in this work.}
  \label{tab:labels}
\end{table}

\section{Logistic Regression Model Training Using Top-k Nearest Examples}
\label{sec:appendixLR}
The classifier in this work is a vanilla logistic regression, similar to probing classifiers, where we have a linear layer on top of contextualized representations \cite{amini2023incontextprobingbuildingrobust}.

We train a logistic regression model on ICL examples using their text and labels. The examples are first converted into numerical representations using a TfidfVectorizer vectorizer, which transforms each sentence into a high-dimensional feature vector. 
An LR model is trained only on these top-$k$ examples as features with their labels. The same vectorizer is used to transform the test prompt into the same feature space. 

\section{Example Prompt Used for Text Classification in ICL}
\label{sec:appendixC}

The following prompt is used in the case of GPT-4o-mini for prompt engineering corresponding to AG News.

\begin{verbatim}
{"role": "system", 
"content": f"Example: We know that 
the classification for the text 
'{examples[i]}', we have
answer: '{labels[i]}'."}

{"role": "system", 
"content": f"According to the 
above provided examples, classify 
the following text. Answer as 
World, Sports, Business, Sci/Tech 
with no explanation."}

{"role": "user", "content": text}
\end{verbatim}

In the given prompt, we first provide a set of example–label pairs determined by the value of $k$ based on their similarity scores ($top_k$ in Algorithm~\ref{alg:cap}). In the next part of the prompt, the model is asked to generate the label for the provided test text.

\section{Relevance Score Labeling of Examples}
\label{sec:relevance}

We mark the top-$k$ high-quality examples with a Boolean label for each example to indicate its relevance. We assign a value of `$1$' to the examples that are relevant to the input test data prompt, and `$0$' to those that are not. Table~\ref{tab:examples} shows two cases where one dataset contains more relevant examples than the other.

\setlength{\tabcolsep}{6pt}
\renewcommand{\arraystretch}{1.2}
\begin{table*}[!t]
  \fontsize{8pt}{8pt}\selectfont
  \begin{tabular}{>{\hspace{0pt}}m{0.1\linewidth}>{\hspace{0pt}}m{0.22\linewidth}>{\hspace{0pt}}m{0.5\linewidth}>{\hspace{0pt}}m{0.08\linewidth}} 
    \toprule
    \textbf{Dataset} & \textbf{Test Data Point} & \textbf{Nearest Neighbors} & \textbf{Relevance Score(0/1)}\\
    \midrule
    \multirow{10}{*}{AG News} 
    & \multirow{10}{*}{Airlines Agree to Cuts at O'Hare} & 
    -- Airlines Agree to Cuts at O'Hare & 1\\
    &   & -- Airlines to Cut Flights at Chicago O'Hare & 1\\
    &   & -- US airlines agree to cut flights at Chicago \#39;s O \#39;Hare & 1\\
    &   & -- 2 Big Carriers at O \#39;Hare to Cut Flights & 1\\
    &   & -- 2 Big Carriers at O \#39;Hare to Cut Flights & 1\\
    &   & -- UPDATE 3-US airlines agree to cut flights at Chicago \#39;s O \#39;Hare & 1\\
    &   & -- Airlines agree to limit O \#39;Hare arrivals & 1\\
    &   & -- FAA: Flight-Reduction Deal Set for O \#39;Hare & 1\\
    &   & -- O \#39;Hare to reduce flight arrivals & 1\\
    &   & -- More Job Cuts Likely At American Airlines & 0\\
    \midrule
    \multirow{10}{*}{\parbox{2.5cm}{SemEval-2019\\(ApexLegends)}} &
    \multirow{10}{*}{\parbox{3.5cm}{Bout to fuck around and stream.\\[0.8ex]
    @PlayApex grind and @Brawlhalla afterwards.\\[0.8ex]
    \texttt{mixer.com/ShinobiSZN} \#BitGang\\[0.8ex]
    \texttt{https://t.co/aBUpN6yjfV}}} &
    -- My goal is just a funny man lmaoo   & 0\\
    &   & -- said My aim is just ridiculous man lmaoo   & 0\\
    &   & -- My aim is just be man lmaoo  & 0 \\
    &   & -- My Wife is just ridiculous man lmaoo   & 0 \\
    &   & -- @PlayApex Past 10 days of pub matchmaking has been a nightmare. Not sure what changed but noticably worse team mates.  & 0  \\
    &   & -- @PlayApex servers are expanding. \u200d.  & 0  \\
    &   & -- @PlayApex servers are whack. The [UNK]..  & 0 \\
    &   & -- . . Stupid robot. pic.twitter.com/nX5ZFPM85G  & 0  \\
    &   &  -- @PlayApex so you guys have zero humans on your life? Best friends account was CLEARLY hacked for illegal purpose of aimbotting/hacking in game, he sent 2 tickets and never got a human response. He has proof, those guys won't even look at it. We have supported your business since day  & 0 \\
    &   & -- @ PlayApex so you guys don't have people in customer support? The best friends account was CLEARLY hacked to hack / hack into the game, it sent 2 tickets and never got a human response.
     & 0\\
    \bottomrule
  \end{tabular}
  \caption{\label{tab:examples}
  \textcolor{red}{WARNING: This table contains language that might be offensive to some.}
    $10$ examples corresponding to a given test data point from two datasets. The first row indicates a test data point associated with high-quality examples (relevance score: $9/10$), whereas the second row presents a test data point with low-quality examples (relevance score: $0/10$).}
\end{table*}

\section{Additional Contingency Matrices for Unweighted kNN}
\label{sec:addcontingency}

As discussed in the paper, we have generated contingency matrices to validate the classification of labels for unweighted-kNN and LR with respect to three LLMs (refer to Section~\ref{sec:contingency}). Figures~\ref{fig:autext} to~\ref{fig:twitter} show the contingency matrices for all other datasets. As previously noted, for several datasets, we observe that the classifiers closely align with the classifications produced by LLMs, with only minor differences as $k$ increases.

By analyzing the figures, we observe that, except for AG News (Figure~\ref{fig:agnews}) and News Category (Figures~\ref{fig:newscat}), the other datasets exhibit lower diagonal dominance. This is likely because the performance of the classifiers heavily depends on the examples, and on such datasets, the classifiers may be influenced by the way the examples are labeled.

\begin{figure}[!h]
 \centering
        \includegraphics[width=0.18\textwidth]{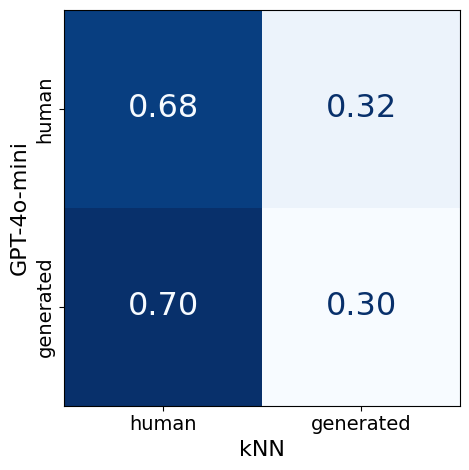}
        \includegraphics[width=0.18\textwidth]{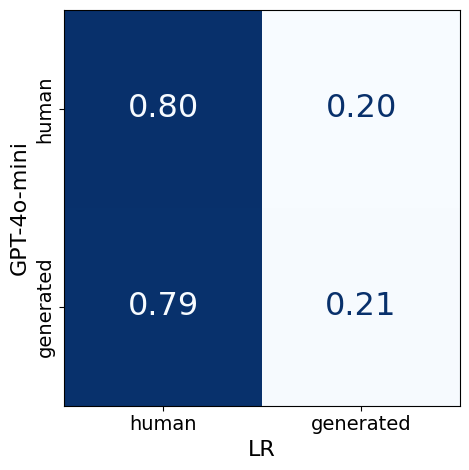}
        \par\smallskip
        \includegraphics[width=0.18\textwidth]{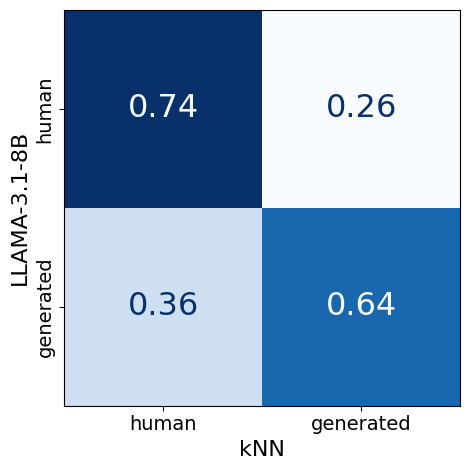}
        \includegraphics[width=0.18\textwidth]{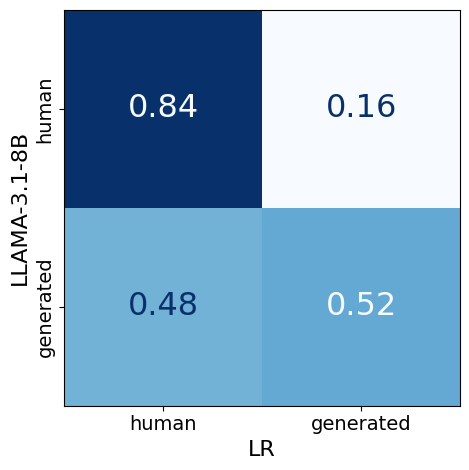}
        \par\smallskip
        \includegraphics[width=0.18\textwidth]{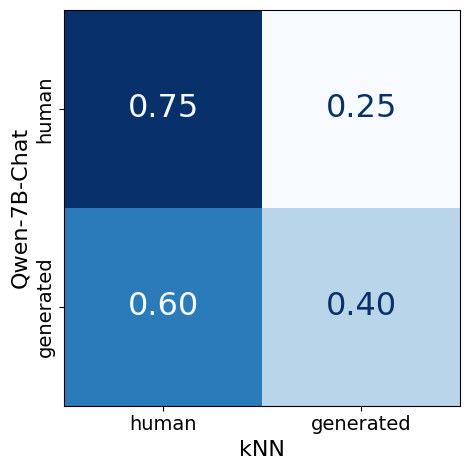}
        \includegraphics[width=0.18\textwidth]{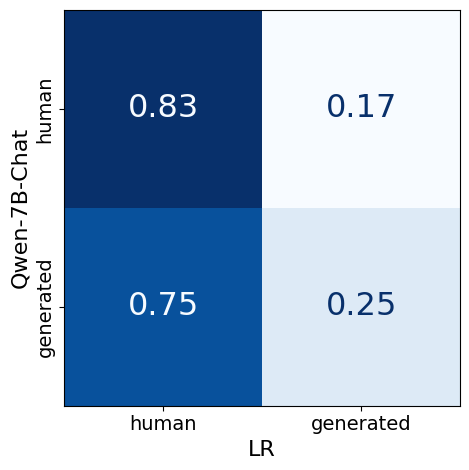}
        \caption{Contingency matrices for AuTexTification: The same configuration follows from Figure~\ref{fig:agnews}.}
  \label{fig:autext}
\end{figure}

\begin{figure}[!h]
 \centering
        \includegraphics[width=0.18\textwidth]{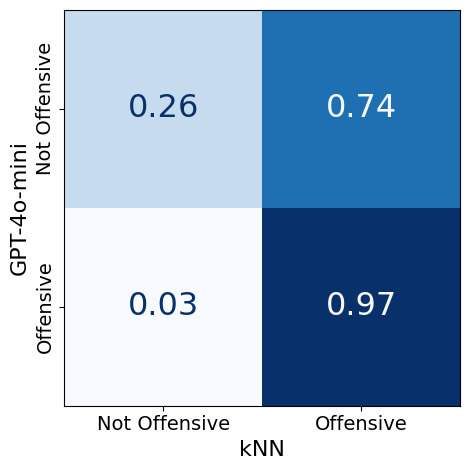}
        \includegraphics[width=0.18\textwidth]{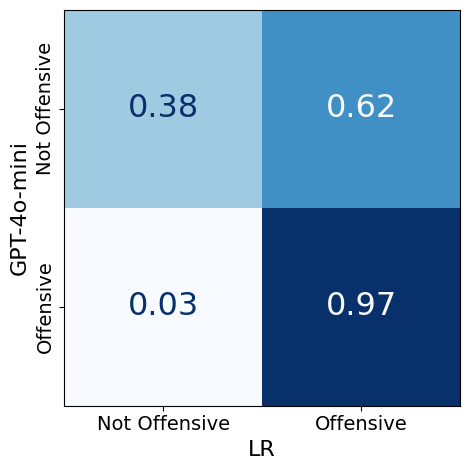}
        \par\smallskip
        \includegraphics[width=0.18\textwidth]{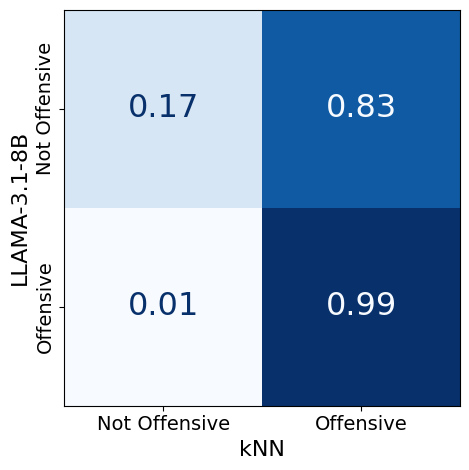}
        \includegraphics[width=0.18\textwidth]{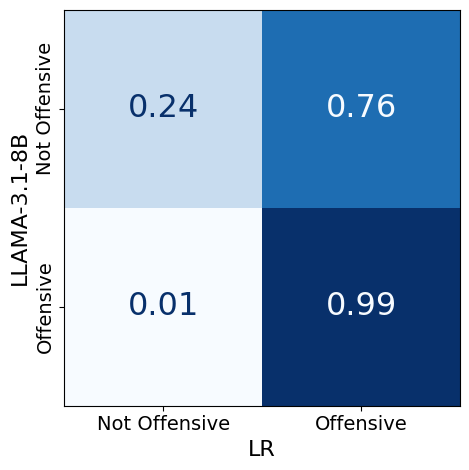}
        \par\smallskip
        \includegraphics[width=0.18\textwidth]{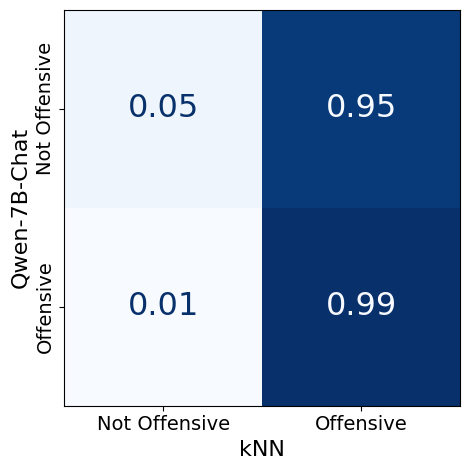}
        \includegraphics[width=0.18\textwidth]{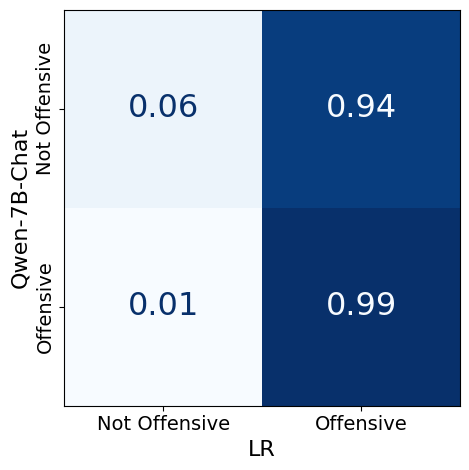}
  \caption{Contingency matrices for Implicit Prompts: The same configuration follows from Figure~\ref{fig:agnews}.}
  \label{fig:implicit}
\end{figure}

\begin{figure}[!h]
 \centering
        \includegraphics[width=0.18\textwidth]{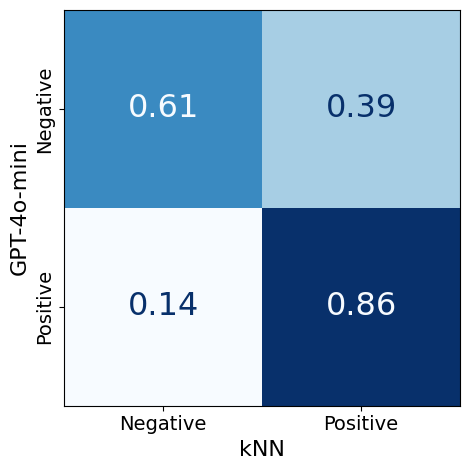}
        \includegraphics[width=0.18\textwidth]{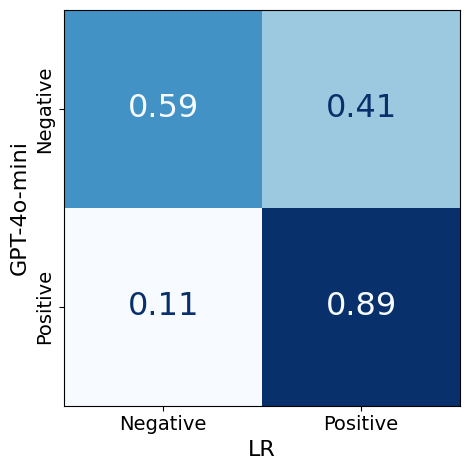}
        \par\smallskip
        \includegraphics[width=0.18\textwidth]{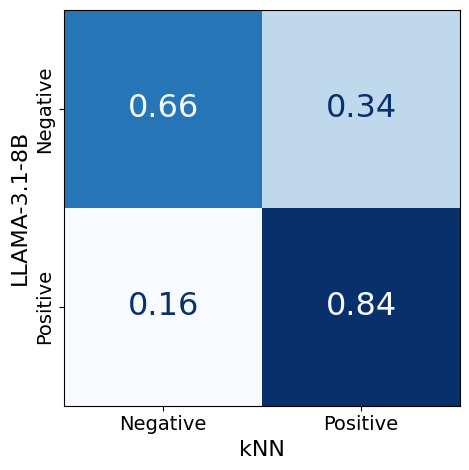}
        \includegraphics[width=0.18\textwidth]{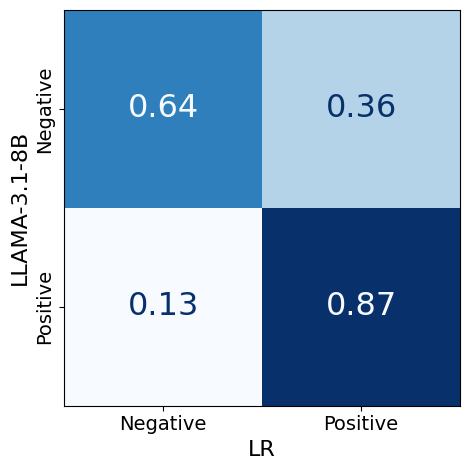}
        \par\smallskip
        \includegraphics[width=0.18\textwidth]{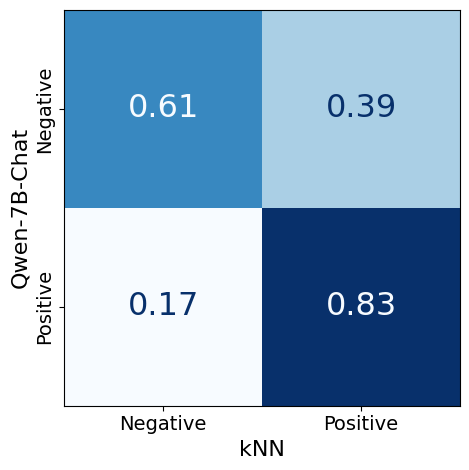}
        \includegraphics[width=0.18\textwidth]{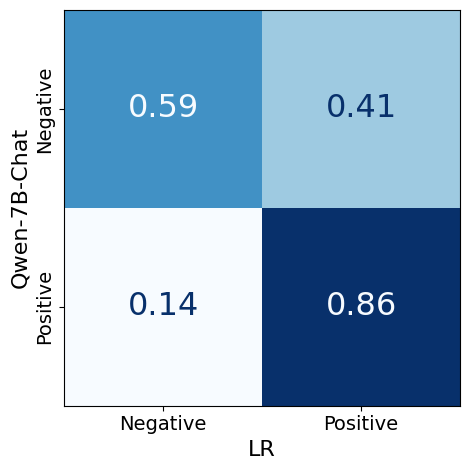}
        \caption{Contingency matrices for SST-2: The same configuration follows from Figure~\ref{fig:agnews}.}
  \label{fig:sst2}
\end{figure}

\begin{figure}[!h]
 \centering
        \includegraphics[width=0.18\textwidth]{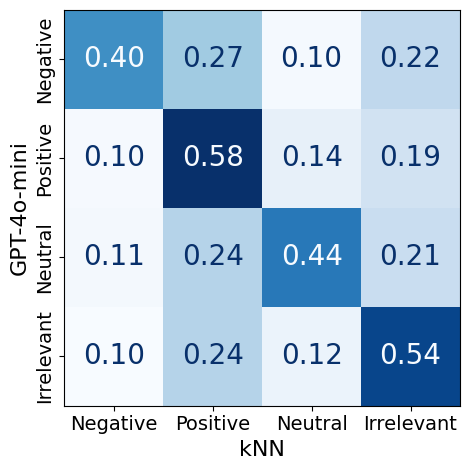}
        \includegraphics[width=0.18\textwidth]{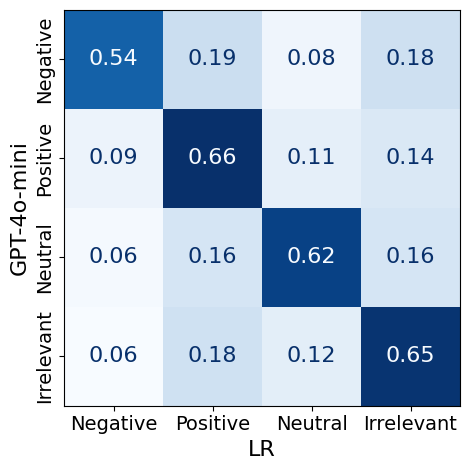}
        \par\smallskip
        \includegraphics[width=0.18\textwidth]{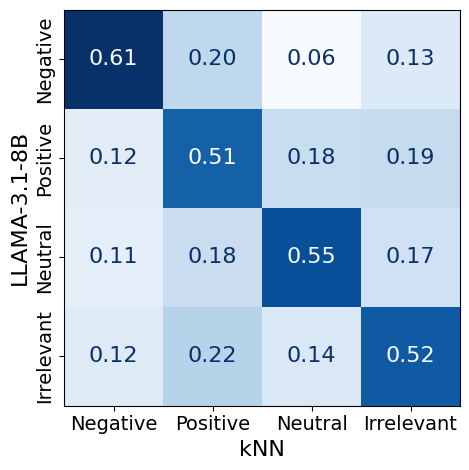}
        \includegraphics[width=0.18\textwidth]{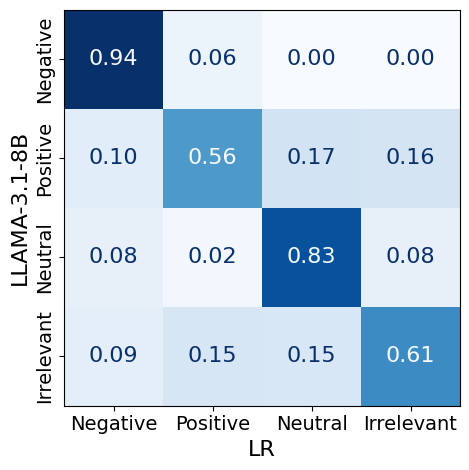}
        \par\smallskip
        \includegraphics[width=0.18\textwidth]{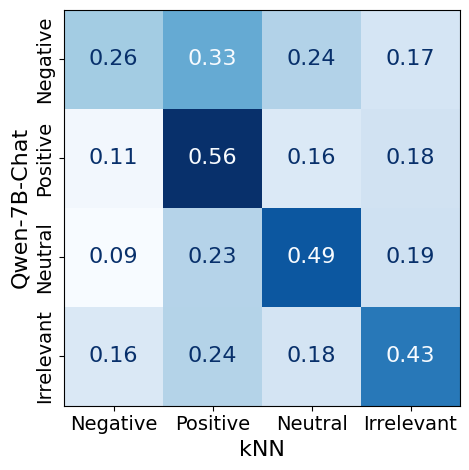}
        \includegraphics[width=0.18\textwidth]{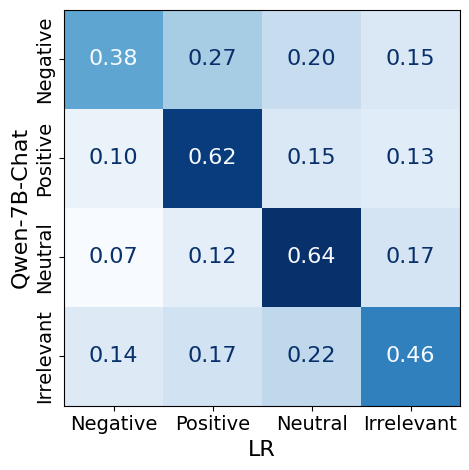}
        \caption{Contingency matrices for SemEval-2019: The same configuration follows from Figure~\ref{fig:agnews}.}
  \label{fig:twitter}
  \vspace{-4mm}
\end{figure}

This is evident because, although the performance metrics of the classifiers are comparable to those of LLMs in the case of AuTexTification (as shown in Table~\ref{accuracy}), the contingency matrices reveal inconsistencies in label classification (Figure~\ref{fig:autext}).
Similarly, in the Implicit Prompts (Figure~\ref{fig:implicit}), the dataset is designed to contain implicit offensive prompts, which means most examples are labeled as `Offensive’. Consequently, the classifiers exhibit strong alignment with these labels.

The SST-2 (Figure~\ref{fig:sst2}) and SemEval-2019 (Figure~\ref{fig:twitter}) datasets show moderate diagonal dominance, meaning that correct predictions are somewhat more frequent along the main diagonal of the confusion matrix. For these datasets, classifiers generally perform worse than LLMs, as indicated in Table~\ref{accuracy}.

Overall, these analyses suggest that classifiers and LLMs overlap in both performance and behavior in some cases, but not all.

\section{Weighted kNN Performance Evaluation}
\label{sec:weightedknn}

\setlength{\tabcolsep}{2.6pt} 
\renewcommand{\arraystretch}{1.7} 
\begin{table*}[!t]
  \centering
  \fontsize{8.2pt}{8pt}\selectfont
  \begin{tabular}{l>{\columncolor{blue!5}}c
  >{\columncolor{blue!5}}c
  >{\columncolor{blue!5}}c
  >{\columncolor{blue!5}}c
  >{\columncolor{yellow!10}}c
  >{\columncolor{yellow!10}}c
  >{\columncolor{yellow!10}}c
  >{\columncolor{yellow!10}}c
  >{\columncolor{red!8}}c
  >{\columncolor{red!8}}c
  >{\columncolor{red!8}}c
  >{\columncolor{red!8}}c  
  >{\columncolor{green!8}}c
  >{\columncolor{green!8}}c
  >{\columncolor{green!8}}c
  >{\columncolor{green!8}}c}
    \toprule
    \multirow{2}{*}{\textbf{Dataset}} 
      & \multicolumn{4}{>{\columncolor{blue!5}}c}{$k=1$} 
      & \multicolumn{4}{>{\columncolor{yellow!10}}c}{$k=10$} 
      & \multicolumn{4}{>{\columncolor{red!8}}c}{$k=20$} 
      & \multicolumn{4}{>{\columncolor{green!8}}c}{$k=30$} \\
    \cmidrule(lr){2-5}
    \cmidrule(lr){6-9}
    \cmidrule(lr){10-13}
    \cmidrule(lr){14-17}
      & kNN & GPT & Llama & Qwen & kNN & GPT & Llama & Qwen & kNN & GPT & Llama & Qwen & kNN & GPT & Llama & Qwen \\
    \midrule
    AG News & 0.85 & 0.86 & 0.82 & 0.81 & 0.88 & 0.89 & 0.88 & 0.88 & 0.88 & 0.89 & 0.88 & 0.88 & 0.87 & 0.90 & 0.87 & 0.87\\
    News Category & 0.80 & 0.86 & 0.80 & 0.77 & 0.83 & 0.88 & 0.85 & 0.81 & 0.83 & 0.88 & 0.85 & 0.81 & 0.83 & 0.88 & 0.85 & 0.80\\
    AuTexTification & 0.54 & 0.55 & 0.54 & 0.53 & 0.59 & 0.57 & 0.60 & 0.54 & 0.59 & 0.57 & 0.60 & 0.54 & 0.58 & 0.57 & 0.58 & 0.54\\
    Implicit Prompts & 0.69 & 0.72 & 0.69 & 0.70 & 0.78 & 0.77 & 0.75 & 0.71 & 0.78 & 0.77 & 0.75 & 0.71 & 0.79 & 0.77 & 0.75 & 0.71\\
    SST-2 & 0.68 & 0.93 & 0.87 & 0.91 & 0.73 & 0.94 & 0.91 & 0.93 & 0.73 & 0.94 & 0.91 & 0.93 & 0.72 & 0.95 & 0.92 & 0.93\\  
    SemEval-2019 & 0.39 & 0.58 & 0.54 & 0.46 & 0.49 & 0.73 & 0.65 & 0.56 & 0.49 & 0.73 & 0.65 & 0.56 & 0.49 & 0.73 & 0.64 & 0.55\\
    \bottomrule
  \end{tabular}
  \captionof{table}{\label{accuracy-weighted}
    Test accuracy on six datasets for weighted-kNN and three LLMs with weighted ICL, using 1, 10, 20, and 30 examples.
  }
  \vspace{-3mm}
\end{table*}

A reasonable question that arises from the results in Section~\ref{compare-models} is: \textit{Why not compare model performance when the models are informed about the example similarity scores?} To explore this, we aim to examine how similar the behavior of ICL would be to weighted kNN if the ICL prompt also includes information about the similarity scores of examples.\footnote{For simplicity, we call the scenario as \textit{weighted ICL}, where we add the similarity scores of each example to the prompt of the original ICL, given in Appendix~\ref{sec:appendixC}.} This analysis is built on the idea of conducting a comparative evaluation between weighted ICL and weighted kNN.\footnote{Definition of weighted kNN is discussed in~\ref{knn}.}

\begin{figure}[!b]
 \centering
        \includegraphics[width=0.18\textwidth]{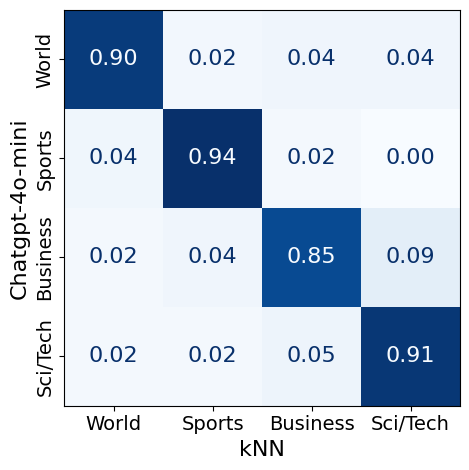}
        \par\smallskip
        \includegraphics[width=0.18\textwidth]{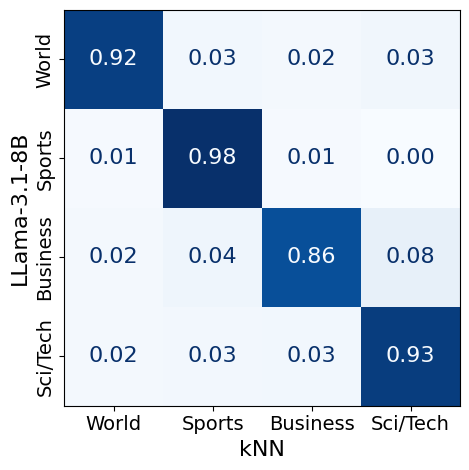}
        \par\smallskip
        \includegraphics[width=0.18\textwidth]{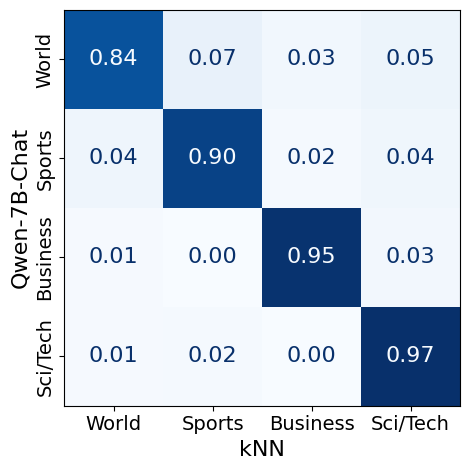}
        \caption{Contingency matrices for AG News: Comparison of weighted kNN vs GPT (top),  weighted kNN vs Llama (middle), and weighted kNN vs Qwen (bottom).}
  \label{fig:weighted-agnews}
\end{figure}

We report the performance values of weighted kNN with three LLMs, considering all datasets for different numbers of $k$ examples in Table~\ref{accuracy-weighted}. As noted earlier in Section~\ref{knn}, the performance is nearly identical to that of the unweighted kNN, suggesting that the inclusion of similarity scores does not substantially influence the results, and the behavior of ICL and kNN remains the same.

In line with the observations from Table~\ref{accuracy}, Table~\ref{accuracy-weighted} indicates that the accuracy of weighted kNN and the weighted ICL on three LLMs is comparable across most cases. Additionally, the overall trend of accuracy values remains consistent with Table~\ref{accuracy}.

To further verify whether the performance of the models is indeed aligned, we examine the contingency matrices between the predictions of each LLM with weighted kNN. Figures~\ref{fig:weighted-agnews} to~\ref{fig:weighted-twitter} present the contingency matrices between weighted kNN (X-axis) and the three LLMs (Y-axis) on the six datasets. 

\begin{figure}[!h]
 \centering
        \includegraphics[width=0.21\textwidth]{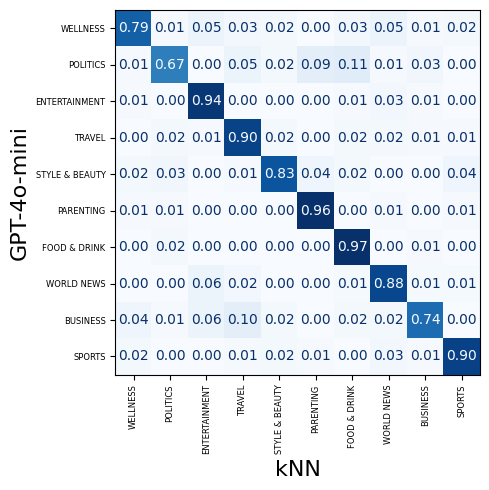}
        \par\smallskip
        \includegraphics[width=0.21\textwidth]{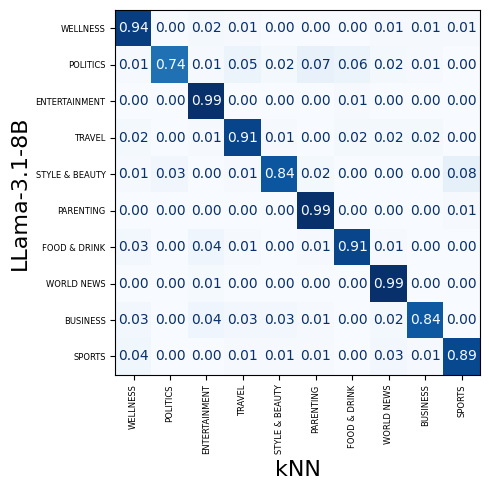}
        \par\smallskip
        \includegraphics[width=0.21\textwidth]{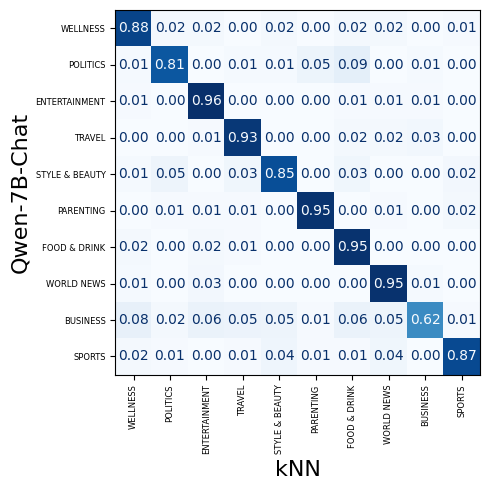}
        \caption{Contingency matrices for News Category: The same configuration follows from Figure~\ref{fig:weighted-agnews}.}
  \label{fig:weighted-news_category}
\end{figure}

\begin{figure}[!h]
 \centering
        \includegraphics[width=0.18\textwidth]{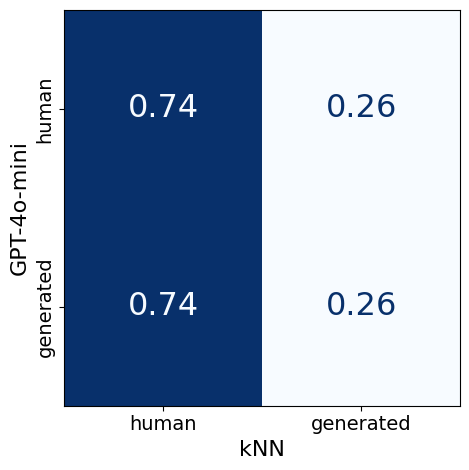}
        \par\smallskip
        \includegraphics[width=0.18\textwidth]{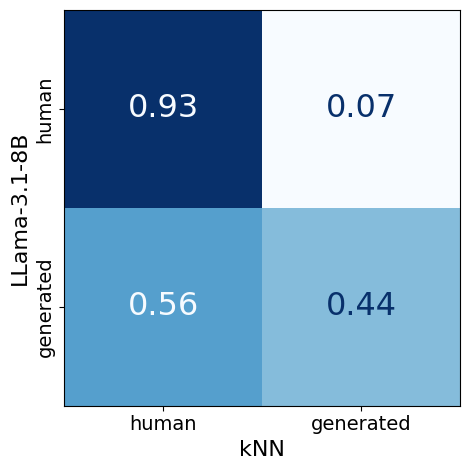}
        \par\smallskip
        \includegraphics[width=0.18\textwidth]{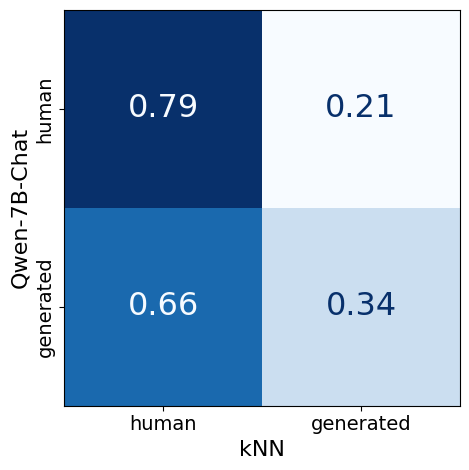}
        \caption{Contingency matrices for AuTexTification: The same configuration follows from Figure~\ref{fig:weighted-agnews}.}
  \label{fig:weighted-autext}
\end{figure}

\begin{figure}[!h]
 \centering
        \includegraphics[width=0.18\textwidth]{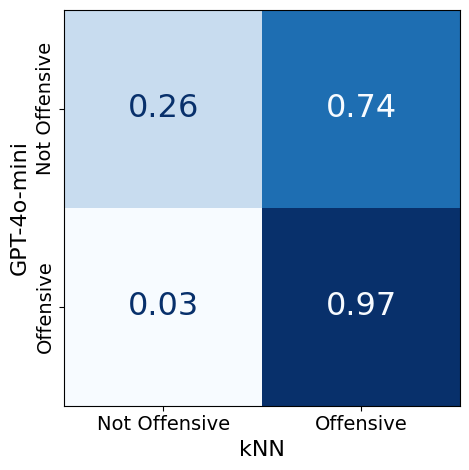}
        \par\smallskip
        \includegraphics[width=0.18\textwidth]{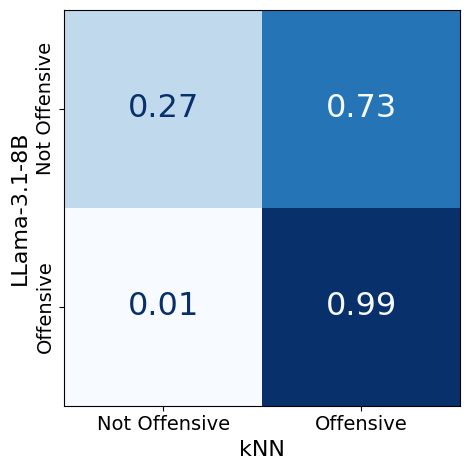}
        \par\smallskip
        \includegraphics[width=0.18\textwidth]{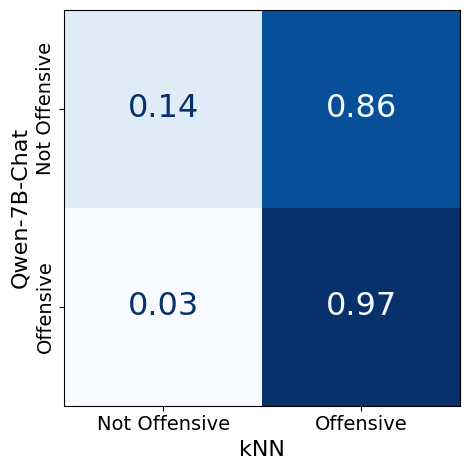}
        \caption{Contingency matrices for Implicit Prompts: The same configuration follows from Figure~\ref{fig:weighted-agnews}.}
  \label{fig:weighted-implicit}
\end{figure}

\begin{figure}[!h]
 \centering
        \includegraphics[width=0.18\textwidth]{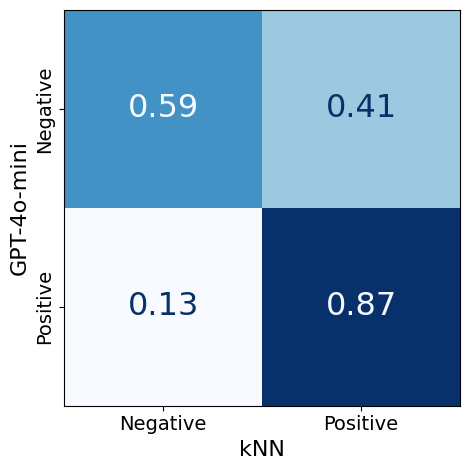}
        \par\smallskip
        \includegraphics[width=0.18\textwidth]{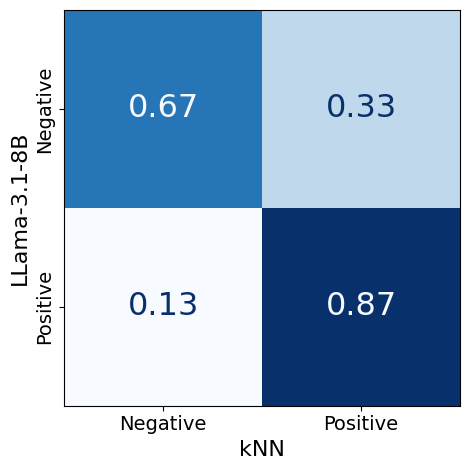}
        \par\smallskip
        \includegraphics[width=0.18\textwidth]{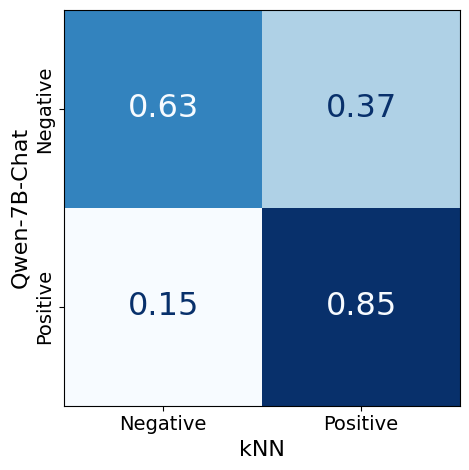}
        \caption{Contingency matrices for SST-2: The same configuration follows from Figure~\ref{fig:weighted-agnews}.}
  \label{fig:weighted-sst2}
\end{figure}

\begin{figure}[!h]
 \centering
        \includegraphics[width=0.18\textwidth]{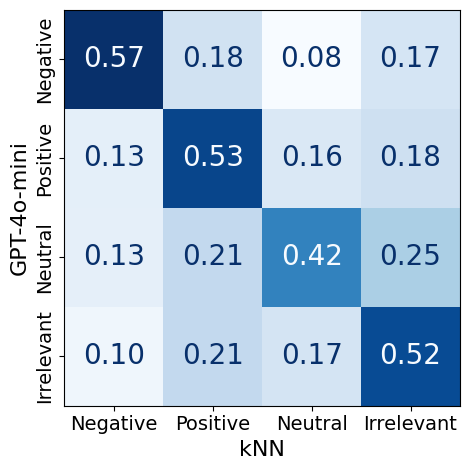}
        \par\smallskip
        \includegraphics[width=0.18\textwidth]{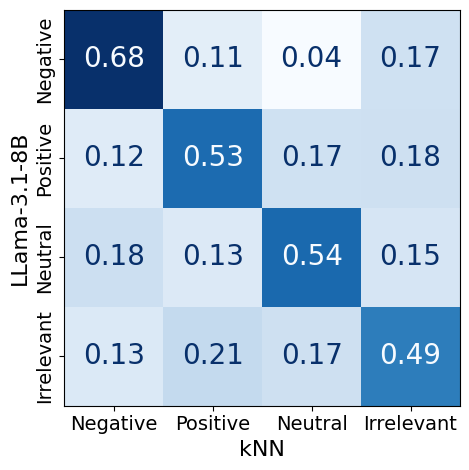}
        \par\smallskip
        \includegraphics[width=0.18\textwidth]{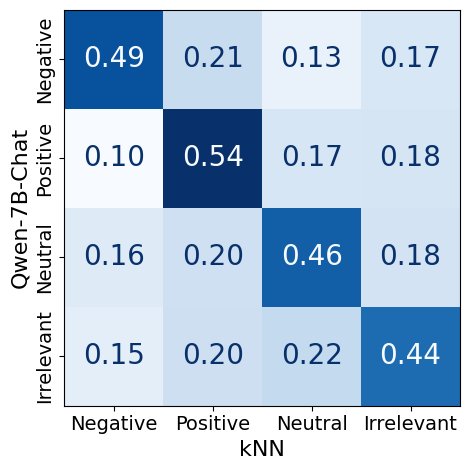}
        \caption{Contingency matrices for SemEval-2019: The same configuration follows from Figure~\ref{fig:weighted-agnews}.}
  \label{fig:weighted-twitter}
\end{figure}

The resulting patterns of contingency matrices for weighted closely resemble those obtained for the unweighted variant of kNN, indicating consistent trends across both kNN configurations.

\section{Does ICL Behave Closer to kNN or LR?}
\label{sec:knnvslr}

The Kappa scores reported in the six tables (\ref{GPT-50}, \ref{gpt-lr-50}, and \ref{llama-50} through~\ref{qwen-lr-50}), which measure behavior similarity between kNN/LR and the LLMs, show small but consistently higher values for kNN. 
For example, for GPT-4o-mini with 
$k=10$, the kNN model achieves a model Kappa (\Kmodel) value of $0.87$ on AG News (first row in Table~\ref{GPT-50}), whereas Logistic Regression (LR) attains a Kappa of $0.84$ (first row in Table~\ref{gpt-lr-50}).
On average, this observation holds for the other datasets and LLMs. 

Further, the correlation values in these tables show a more intuitive behavior for kNN. This is evident from the increasing Pearson ($r$) and $R^2$ values as $k$ grows, e.g., $R^2$ grows from $31.3\%$ to $63.1\%$ from $k=1$ to $k=20$ for $Correlation$-$A$ (Table~\ref{GPT-50}). This trend does not hold for LR: $R^2$ drops from $31.3\%$ to $26.8\%$ in the same scenario (Table~\ref{gpt-lr-50}).

These observations suggest that ICL behaves closer to kNN than LR. However, as we have a very simple gradient-based classifier, the results may vary in larger models. In our setting involving simplified models, we focus on understanding the behavior of ICL with these classifiers.

\setlength{\tabcolsep}{4pt}
\renewcommand{\arraystretch}{1.1}
\begin{table*}[!b]
\fontsize{10pt}{10pt}\selectfont
\centering
  \begin{tabular}{p{3.2cm}ccccccccc}
    \toprule
    \multirow{2}{*}{Datasets \textbackslash{}  Neighbors}    & \multicolumn{3}{c}{$k=1$} & \multicolumn{3}{c}{$k=10$} & \multicolumn{3}{c}{$k=20$}\\
    \cmidrule(lr){2-4}
    \cmidrule(lr){5-7}
    \cmidrule(lr){8-10}
    & \Rhum & \Kmodel & \Rllm  & \Rhum & \Kmodel & \Rllm & \Rhum & \Kmodel & \Rllm\\
    \midrule
     AG News & 0.9 & 0.87 & 1 & 0.794 & 0.84 & 0.976 & 0.7 & 0.84 & 0.952 \\
    News Category & 0.8 & 0.87 & 0.88 & 0.664 & 0.84 & 0.834 & 0.553 & 0.87 & 0.795 \\
    AuTexTification & 0.5 & 0.06 & 0.66 & 0.312 & 0.11 & 0.546 & 0.267 & 0.05 & 0.507 \\
    Implicit Prompts & 0.96 & 0.19 & 0.98 & 0.756 & 0.19 & 0.966 & 0.627 & 0.3 & 0.956 \\
    SST-2 & 0.7 & 0.31 & 0.98 & 0.596 & 0.42 & 0.97 & 0.471 & 0.53 & 0.961 \\
    SemEval-2019 & 0.2 & 0.12 & 0.38 & 0.21 & 0.46 & 0.57 & 0.154 & 0.49 & 0.581   \\
    \midrule
    \rowcolor{lightgreen!45} $Correlation$-$A$ &  
    \multicolumn{3}{c}{\parbox[c]{3.2cm}{\centering $r$: 0.559 \\ [0.7mm] $R^{2}$: 0.313}} & 
    \multicolumn{3}{c}{\parbox[c]{3.2cm}{\centering $r$: 0.431 \\ [0.7mm] $R^{2}$: 0.186}} &  
    \multicolumn{3}{c}{\parbox[c]{3.2cm}{\centering $r$: 0.517  \\ [0.7mm] $R^{2}$: 0.268}} \\
    \midrule
    \rowcolor{lightblue!45} $Correlation$-$B$ &  
    \multicolumn{3}{c}{\parbox[c]{3.2cm}{\centering $r$: 0.528  \\ [0.7mm] $R^{2}$: 0.279}} & 
    \multicolumn{3}{c}{\parbox[c]{3.2cm}{\centering $r$: 0.368  \\ [0.7mm] $R^{2}$: 0.135}} &  
    \multicolumn{3}{c}{\parbox[c]{3.2cm}{\centering $r$: 0.494  \\ [0.7mm] $R^{2}$: 0.244}} \\
    \bottomrule
  \end{tabular}
  \caption{\label{gpt-lr-50}
    Comparison of {\bf GPT-4o-mini} and \textbf{LR}.
    The same configuration follows from Table~\ref{GPT-50}.
  }
\end{table*}

\setlength{\tabcolsep}{4pt}
\renewcommand{\arraystretch}{1.1}
\begin{table*}[!hb]
\fontsize{10pt}{10pt}\selectfont
\centering
  \begin{tabular}{p{3.2cm}ccccccccc}
    \toprule
    \multirow{2}{*}{Datasets \textbackslash{}  Neighbors}    & \multicolumn{3}{c}{$k=1$} & \multicolumn{3}{c}{$k=10$} & \multicolumn{3}{c}{$k=20$}\\
    \cmidrule(lr){2-4}
    \cmidrule(lr){5-7}
    \cmidrule(lr){8-10}
    & \Rhum & \Kmodel & \Rllm  & \Rhum & \Kmodel & \Rllm & \Rhum & \Kmodel & \Rllm\\
    \midrule
     AG News & 0.9 & 0.92 & 0.86 & 0.794 & 1 & 0.722 & 0.7 & 0.89 & 0.696   \\
    News Category & 0.8 & 0.89 & 0.82 & 0.664 & 0.89 & 0.762 & 0.553 & 0.91 & 0.736   \\
    AuTexTification & 0.5 & 1 & 0.8 & 0.312 & 0.67 & 0.764 & 0.267 & 0.35 & 0.749   \\
    Implicit Prompts & 0.96 & 0.29 & 0.8 & 0.756 & 0.52 & 0.672 & 0.627 & 0.21 & 0.625   \\
    SST-2 & 0.7 & 0.51 & 0.92 & 0.596 & 0.74 & 0.884 & 0.471 & 0.7 & 0.868   \\
    SemEval-2019 & 0.2 & 0.15 & 0.48 & 0.21 & 0.31 & 0.594 & 0.154 & 0.32 & 0.594   \\
    \midrule
    \rowcolor{lightgreen!45} $Correlation$-$A$ &  
    \multicolumn{3}{c}{\parbox[c]{3.2cm}{\centering $r$: 0.315 \\ [0.7mm] $R^{2}$: 0.099}} & 
    \multicolumn{3}{c}{\parbox[c]{3.2cm}{\centering $r$: 0.665 \\ [0.7mm] $R^{2}$: 0.442}} &  
    \multicolumn{3}{c}{\parbox[c]{3.2cm}{\centering $r$: 0.516  \\ [0.7mm] $R^{2}$: 0.266}} \\
    \midrule
    \rowcolor{lightblue!45} $Correlation$-$B$ &  
    \multicolumn{3}{c}{\parbox[c]{3.2cm}{\centering $r$: 0.582  \\ [0.7mm] $R^{2}$: 0.339}} & 
    \multicolumn{3}{c}{\parbox[c]{3.2cm}{\centering $r$: 0.594  \\ [0.7mm] $R^{2}$: 0.352}} &  
    \multicolumn{3}{c}{\parbox[c]{3.2cm}{\centering $r$: 0.503  \\ [0.7mm] $R^{2}$: 0.253}} \\
    \bottomrule
  \end{tabular}
  \caption{\label{llama-50}
    Comparison of {\bf Llama-3.1-8B} and \textbf{kNN}.
    The same configuration follows from Table~\ref{GPT-50}.
  }
\end{table*}

\section{Additional Correlation Results}
\label{sec:appendixG}

Extending our analysis from Table~\ref{GPT-50} to the comparison between GPT-4o-mini and LR, we present the correlation values in Table~\ref{gpt-lr-50}. Evidently, we note that $R^{2}$ value for  $Correlation$-$A$ with $k = 20$ indicates that only up to $26.8\%$ of the behavioral similarity can be explained by example relevance. This demonstrates that KNN exhibits a stronger correlation compared to LR.

Additionally, Tables~\ref{llama-50} and~\ref{llama-lr-50} show the correlation results for Llama-3.1-8B with kNN and LR, respectively. As expected, kNN has a comparatively better correlation with the LLM when compared to LR. Similarly, we extend our analysis for QWEN-7B-Chat in Tables~\ref{qwen-50} and~\ref{qwen-lr-50}. kNN tends to perform closer to the LLM in this case as well.

\setlength{\tabcolsep}{4pt}
\renewcommand{\arraystretch}{1.1}
\begin{table*}[!t]
\fontsize{10pt}{10pt}\selectfont
\centering
  \begin{tabular}{p{3.2cm}ccccccccc}
    \toprule
    \multirow{2}{*}{Datasets \textbackslash{}  Neighbors}    & \multicolumn{3}{c}{$k=1$} & \multicolumn{3}{c}{$k=10$} & \multicolumn{3}{c}{$k=20$}\\
    \cmidrule(lr){2-4}
    \cmidrule(lr){5-7}
    \cmidrule(lr){8-10}
    & \Rhum & \Kmodel & \Rllm  & \Rhum & \Kmodel & \Rllm & \Rhum & \Kmodel & \Rllm\\
    \midrule
     AG News &0.9 &0.92 &0.86 &0.794 &0.92 &0.722 &0.7 &0.87 &0.696  \\
    News Category &0.8 &0.89 &0.82 &0.664 &0.89 &0.762 &0.553 &0.93 &0.736   \\
    AuTexTification &0.5 &1 &0.8 &0.312 &0.62 &0.764 &0.267 &0.46 &0.749   \\
    Implicit Prompts &0.96 &0.29 &0.8 &0.756 &0.41 &0.672 &0.627 &0.26 &0.625   \\
    SST-2 &0.7 &0.51 &0.92 &0.596 &0.65 &0.884 &0.471 &0.74 &0.868   \\
    SemEval-2019 &0.2 &0.15 &0.48 &0.21 &0.43 &0.594 &0.154 &0.39 &0.594 \\
    \midrule
    \rowcolor{lightgreen!45} $Correlation$-$A$ &  
    \multicolumn{3}{c}{\parbox[c]{3.2cm}{\centering $r$: 0.315 \\ [0.7mm] $R^{2}$: 0.099}} & 
    \multicolumn{3}{c}{\parbox[c]{3.2cm}{\centering $r$: 0.480 \\ [0.7mm] $R^{2}$: 0.230}} &  
    \multicolumn{3}{c}{\parbox[c]{3.2cm}{\centering $r$: 0.449  \\ [0.7mm] $R^{2}$: 0.202}} \\
    \midrule
    \rowcolor{lightblue!45} $Correlation$-$B$ &  
    \multicolumn{3}{c}{\parbox[c]{3.2cm}{\centering $r$: 0.582  \\ [0.7mm] $R^{2}$: 0.339}} & 
    \multicolumn{3}{c}{\parbox[c]{3.2cm}{\centering $r$: 0.453  \\ [0.7mm] $R^{2}$: 0.206}} &  
    \multicolumn{3}{c}{\parbox[c]{3.2cm}{\centering $r$: 0.552  \\ [0.7mm] $R^{2}$: 0.305}} \\
    \bottomrule
  \end{tabular}
  \caption{\label{llama-lr-50}
    Comparison of {\bf Llama-3.1-8B} and \textbf{LR}.
    The same configuration follows from Table~\ref{GPT-50}.
  }
\end{table*}

\setlength{\tabcolsep}{4pt}
\renewcommand{\arraystretch}{1.1}
\begin{table*}[!t]
\fontsize{10pt}{10pt}\selectfont
\centering
  \begin{tabular}{p{3.2cm}ccccccccc}
    \toprule
    \multirow{2}{*}{Datasets \textbackslash{}  Neighbors}    & \multicolumn{3}{c}{$k=1$} & \multicolumn{3}{c}{$k=10$} & \multicolumn{3}{c}{$k=20$}\\
    \cmidrule(lr){2-4}
    \cmidrule(lr){5-7}
    \cmidrule(lr){8-10}
    & \Rhum & \Kmodel & \Rllm  & \Rhum & \Kmodel & \Rllm & \Rhum & \Kmodel & \Rllm\\
    \midrule
     AG News &0.9 &0.92 &0.82 &0.794 &0.89 &0.688 &0.7 &0.89 &0.629   \\
    News Category &0.8 &0.37 &0.54 &0.664 &0.63 &0.418 &0.553 &0.76 &0.389   \\
    AuTexTification &0.5 &0.02 &0.6 &0.312 &-0.12 &0.586 &0.267 &0.1 &0.549   \\
    Implicit Prompts &0.96 &0.11 &0.8 &0.756 &0.15 &0.694 &0.627 &0.1 &0.646   \\
    SST-2 &0.7 &0.34 &0.68 &0.596 &0.49 &0.766 &0.471 &0.61 &0.762   \\
    SemEval-2019 &0.2 &0.08 &0.42 &0.21 &0.15 &0.474 &0.154 &0.33 &0.476   \\
    \midrule
    \rowcolor{lightgreen!45} $Correlation$-$A$ &  
    \multicolumn{3}{c}{\parbox[c]{3.2cm}{\centering $r$: 0.525 \\ [0.7mm] $R^{2}$: 0.276}} & 
    \multicolumn{3}{c}{\parbox[c]{3.2cm}{\centering $r$: 0.691 \\ [0.7mm] $R^{2}$: 0.477}} &  
    \multicolumn{3}{c}{\parbox[c]{3.2cm}{\centering $r$: 0.506  \\ [0.7mm] $R^{2}$: 0.256}} \\
    \midrule
    \rowcolor{lightblue!45} $Correlation$-$B$ &  
    \multicolumn{3}{c}{\parbox[c]{3.2cm}{\centering $r$: 0.526  \\ [0.7mm] $R^{2}$: 0.277}} & 
    \multicolumn{3}{c}{\parbox[c]{3.2cm}{\centering $r$: 0.128  \\ [0.7mm] $R^{2}$: 0.016}} &  
    \multicolumn{3}{c}{\parbox[c]{3.2cm}{\centering $r$: -0.035  \\ [0.7mm] $R^{2}$: 0.001}} \\
    \bottomrule
  \end{tabular}
  \caption{\label{qwen-50}
    Comparison of {\bf QWEN-7B-Chat} and \textbf{kNN}.
    The same configuration follows from Table~\ref{GPT-50}.
  }
\end{table*}

\setlength{\tabcolsep}{4pt}
\renewcommand{\arraystretch}{1.1}
\begin{table*}[!t]
\fontsize{10pt}{10pt}\selectfont
\centering
  \begin{tabular}{p{3.2cm}ccccccccc}
    \toprule
    \multirow{2}{*}{Datasets \textbackslash{}  Neighbors}    & \multicolumn{3}{c}{$k=1$} & \multicolumn{3}{c}{$k=10$} & \multicolumn{3}{c}{$k=20$}\\
    \cmidrule(lr){2-4}
    \cmidrule(lr){5-7}
    \cmidrule(lr){8-10}
    & \Rhum & \Kmodel & \Rllm  & \Rhum & \Kmodel & \Rllm & \Rhum & \Kmodel & \Rllm\\
    \midrule
     AG News & 0.9 & 0.92 & 0.82 & 0.794 & 0.87 & 0.688 & 0.7 & 0.87 & 0.629 \\
    News Category & 0.8 & 0.37 & 0.54 & 0.664 & 0.61 & 0.418 & 0.553 & 0.76 & 0.389 \\
    AuTexTification & 0.5 & 0.02 & 0.6 & 0.312 & -0.14 & 0.586 & 0.267 & 0.02 & 0.549 \\
    Implicit Prompts & 0.96 & 0.11 & 0.8 & 0.756 & 0.11 & 0.694 & 0.627 & 0.06 & 0.646 \\
    SST-2 & 0.7 & 0.34 & 0.68 & 0.596 & 0.42 & 0.766 & 0.471 & 0.57 & 0.762 \\
    SemEval-2019 & 0.2 & 0.08 & 0.42 & 0.21 & 0.21 & 0.474 & 0.154 & 0.38 & 0.476 \\
    \midrule
    \rowcolor{lightgreen!45} $Correlation$-$A$ &  
    \multicolumn{3}{c}{\parbox[c]{3.2cm}{\centering $r$: 0.525 \\ [0.7mm] $R^{2}$: 0.276}} & 
    \multicolumn{3}{c}{\parbox[c]{3.2cm}{\centering $r$: 0.629 \\ [0.7mm] $R^{2}$: 0.395}} &  
    \multicolumn{3}{c}{\parbox[c]{3.2cm}{\centering $r$: 0.453  \\ [0.7mm] $R^{2}$: 0.205}} \\
    \midrule
    \rowcolor{lightblue!45} $Correlation$-$B$ &  
    \multicolumn{3}{c}{\parbox[c]{3.2cm}{\centering $r$: 0.526  \\ [0.7mm] $R^{2}$: 0.277}} & 
    \multicolumn{3}{c}{\parbox[c]{3.2cm}{\centering $r$: 0.049  \\ [0.7mm] $R^{2}$: 0.002}} &  
    \multicolumn{3}{c}{\parbox[c]{3.2cm}{\centering $r$: -0.094  \\ [0.7mm] $R^{2}$: 0.009}} \\
    \bottomrule
  \end{tabular}
  \caption{\label{qwen-lr-50}
    Comparison of {\bf QWEN-7B-Chat} and \textbf{LR}.
    The same configuration follows from Table~\ref{GPT-50}.
  }
\end{table*}

In summary, the observations from Section~\ref{sec:correlation} generally hold for the other two LLMs, with a couple of caveats. First, the correlations are weaker for $k=1$, suggesting that one ICL example is not sufficient to influence the predictions of the two LLMs. Second, the correlation values increase considerably from $k=1$ to $k=10$, but not always from $k=10$ to $k=20$. This suggests that these two LLMs can become ``confused'' by too many ICL examples. This is reflected in Table~\ref{accuracy} where there is a decrease in the accuracy value of Qwen on SemEval-2019 from $k=10$ to $k=20$. 

Additionally, Appendix~\ref{sec:poorvszero} highlights that the LLM indeed backs off to its parametric memory when using low-quality demonstrations in ICL.This was validated by comparing the outputs of the LLM with poor-quality examples with a zero-shot configuration of the same LLM.

\section{Empirical Validation that LLMs under ICL with Poor-Quality Examples Back off to Parametric Memory}
\label{sec:poorvszero}

To validate the hypothesis that LLMs, when using ICL with poor-quality examples, back off to parametric memory, we compare their outputs with the same LLM in a zero-shot configuration. If our hypothesis is correct, both settings should produce similar labels, with a high Kappa agreement between them.
In particular, we ran the three LLMs for $50$ test samples under two settings: one with no examples (zero-shot) and the other with $10$ poor examples. Table~\ref{tab:zero-poor} shows the Cohen's Kappa agreement between the two settings for different LLMs.

\setlength{\tabcolsep}{6.7pt}
\renewcommand{\arraystretch}{1.2}
\begin{table}[!h]
\fontsize{10pt}{10pt}\selectfont
  \begin{tabular}{p{24mm}
  >{\centering\arraybackslash}p{11mm}
  >{\centering\arraybackslash}p{11mm}
  >{\centering\arraybackslash}p{13mm}
}
    \toprule
    Datasets & Gpt-4o-mini & Llama-3.1-8B & QWEN-7B-Chat \\
    \midrule
    AG News & 0.86 & 0.57 & 0.45 \\
    News Category & 0.98 & 0.43 & 0.63\\
    AuTexTification & 0.96 & 0.09 & 0.05\\
    Implicit Prompts & 0.69 & 0.59 & 0.40\\
    SST-2  & 0.60 & 0.68 & 0.96\\
    SemEval-2019 & 0.13 & 0.45 & 0.42\\
    \bottomrule
  \end{tabular}
  \caption{Kappa agreement between zero-shot and poor-quality examples for all datasets.}
  \label{tab:zero-poor}
\end{table}

The vast majority of Kappa values in Table~\ref{tab:zero-poor} indicate substantial to near-perfect agreement, which aligns with our expectation that LLM relies on parametric memory when example quality is low, resulting in behavior comparable to the zero-shot. 
This is an encouraging result, which suggests that LLMs can ignore low-quality examples. 
There are some exceptions to the rule, which suggests that in some situations, low-quality demonstrations influence LLM behavior, like the SemEval-2019 dataset on Gpt-4o-mini, and AuTexTification on Llama-3.1-8B and QWEN-7B-Chat.

\section{Selected Examples Do not Necessarily Share the Same Label}
\label{sec:appendixE}

A potential concern from our observations is that the similar behavior of LLM-based classifiers might be because of the possibility that selected examples in ICL or kNN\footnote{We report results obtained using an unweighted kNN approach in this section.} have all the same label. However, as shown below, this largely does not apply to our scenario.
To verify this, we analyze the proportion of test data points in each dataset whose selected examples have the same versus differing labels (see Table~\ref{tab:same-label-stats}).

\setlength{\tabcolsep}{7pt}
\renewcommand{\arraystretch}{1.2}
\begin{table}[!h]
\fontsize{10pt}{10pt}\selectfont
  \begin{tabular}{p{24mm}
  >{\centering\arraybackslash}p{11mm}
  >{\centering\arraybackslash}p{11mm}
  >{\centering\arraybackslash}p{11mm}
}
    \toprule
    Datasets & $k=10$ & $k=20$ & $k=30$ \\
    \midrule
    AG News & 49.35 & 33.37 & 23.676 \\
    News Category & 34 & 18.8 & 10.8\\
    AuTexTification & 4.6 & 0.2 & 0\\
    Implicit Prompts & 15.95 & 3.987 & 0.42\\
    SST-2 & 26.49 & 11.353 & 5.73\\
    SemEval-2019 & 5.2 & 0.6 & 0.2\\
    \bottomrule
  \end{tabular}
  \caption{Percentage of datapoints where all the examples have the same label for increasing $k$.}
  \label{tab:same-label-stats}
\end{table}

In each of these cases, we further check the percentage of datapoints where the labels predicted by kNN and LLM align. For example, in Table~\ref{tab:gpt-same-label-align-stats}, for the $49.35\%$ of datapoints where AG News has all the same labels (from Table~\ref{tab:same-label-stats}), kNN and Gpt-4o-mini make the same prediction $99.4\%$ of the time. Similarly, as shown in Table~\ref{tab:gpt-diff-label-align-stats}, among the $50.65\%$ of data points where AG News has different labels, kNN and Gpt-4o-mini agree in their predictions $81.85\%$ of the time.

\setlength{\tabcolsep}{7pt}
\renewcommand{\arraystretch}{1.2}
\begin{table}[!h]
\fontsize{10pt}{10pt}\selectfont
  \begin{tabular}{p{24mm}
  >{\centering\arraybackslash}p{11mm}
  >{\centering\arraybackslash}p{11mm}
  >{\centering\arraybackslash}p{11mm}
}
    \toprule
    Datasets & $k=10$ & $k=20$ & $k=30$ \\
    \midrule
    AG News & 99.4 & 99.7 & 100 \\
    News Category & 100 & 100 & 100\\
    AuTexTification  & 84.78 & 100 & 0\\
    Implicit Prompts  & 98.86 & 100 & 100\\
    SST-2  & 87.45 & 94.95 & 96\\
    SemEval-2019  & 69.23 & 83.33 & 100\\
    \bottomrule
  \end{tabular}
  \caption{Percentage of datapoints out of the ones that have the \textit{same} labels, where the prediction of kNN and Gpt-4o-mini match.}
  \label{tab:gpt-same-label-align-stats}
\end{table}

\setlength{\tabcolsep}{7pt}
\renewcommand{\arraystretch}{1.2}
\begin{table}[!h]
\fontsize{10pt}{10pt}\selectfont
  \begin{tabular}{p{24mm}
  >{\centering\arraybackslash}p{11mm}
  >{\centering\arraybackslash}p{11mm}
  >{\centering\arraybackslash}p{11mm}
}
    \toprule
    Datasets & $k=10$ & $k=20$ & $k=30$ \\
    \midrule
    AG News & 81.85 & 84.56 & 85.34 \\
    News Category  & 75.53 & 79.66 & 81.03\\
    AuTexTification  & 54.61 & 53.61 & 55.4\\
    Implicit Prompts  & 91.52 & 92.7 & 93.51\\
    SST-2  & 65.83 & 69.47 & 70.92\\
    SemEval-2019  & 47.58 & 49.8 & 46.5\\
    \bottomrule
  \end{tabular}
  \caption{Percentage of datapoints out of the ones that have \textit{different} labels, where the prediction of kNN and Gpt-4o-mini match.}
  \label{tab:gpt-diff-label-align-stats}
\end{table}

While we expect the predictions to align well with the labels of examples in the first scenario, the second scenario also shows a high level of agreement.
This supports our argument, as kNN and Gpt-4o-mini tend to make similar predictions even when the example labels are not identical.

\setlength{\tabcolsep}{7pt}
\renewcommand{\arraystretch}{1.2}
\begin{table}[!t]
\fontsize{10pt}{10pt}\selectfont
  \begin{tabular}{p{24mm}
  >{\centering\arraybackslash}p{11mm}
  >{\centering\arraybackslash}p{11mm}
  >{\centering\arraybackslash}p{11mm}
}
    \toprule
    Datasets & $k=10$ & $k=20$ & $k=30$ \\
    \midrule
    AG News & 100 & 100 & 100 \\
    News Category & 100  & 100 & 100 \\
    AuTexTification  & 100 & 100 & 0 \\
    Implicit Prompts  & 95.83 & 98.48 & 100 \\
    SST-2 & 95.23 & 96.97  & 98.0 \\
    SemEval-2019  & 71.15 & 100 & 100 \\
    \bottomrule
  \end{tabular}
  \caption{Percentage of datapoints out of the ones that have the \textit{same} labels, where the prediction of kNN and Llama-3.1-8B match.}
  \label{tab:llama-same-label-align-stats}
  \vspace{-1mm}
\end{table}

\setlength{\tabcolsep}{7pt}
\renewcommand{\arraystretch}{1.2}
\begin{table}[!t]
\fontsize{10pt}{10pt}\selectfont
  \begin{tabular}{p{24mm}
  >{\centering\arraybackslash}p{11mm}
  >{\centering\arraybackslash}p{11mm}
  >{\centering\arraybackslash}p{11mm}
}
    \toprule
    Datasets & $k=10$ & $k=20$ & $k=30$ \\
    \midrule
    AG News & 83.83 & 87.26 & 88.09 \\
    News Category & 84.44 & 86.64 & 86.83\\
    AuTexTification  & 68.44 & 72.44 & 73.3 \\
    Implicit Prompts  & 78.15 & 83.57 & 83.43\\
    SST-2 & 71.45 & 72.70 & 75.18 \\
    SemEval-2019  & 48.74 & 51.91 & 51.6 \\
    \bottomrule
  \end{tabular}
  \caption{Percentage of datapoints out of the ones that have the \textit{different} labels, where the prediction of kNN and Llama-3.1-8B match.}
  \label{tab:llama-diff-label-align-stats}
  \vspace{-1mm}
\end{table}

\setlength{\tabcolsep}{7pt}
\renewcommand{\arraystretch}{1.2}
\begin{table}[!t]
\fontsize{10pt}{10pt}\selectfont
  \begin{tabular}{p{24mm}
  >{\centering\arraybackslash}p{11mm}
  >{\centering\arraybackslash}p{11mm}
  >{\centering\arraybackslash}p{11mm}
}
    \toprule
    Datasets & $k=10$ & $k=20$ & $k=30$ \\
    \midrule
    AG News & 100 & 100 & 100 \\
    News Category & 81.87 & 86.17 & 93.52 \\
    AuTexTification  & 86.95 & 100 & 0 \\
    Implicit Prompts  & 35.60 & 27.27 & 14.28 \\
    SST-2 & 85.71 & 95.96  & 96\\
    SemEval-2019  & 55.77 & 66.67 &  100\\
    \bottomrule
  \end{tabular}
  \caption{Percentage of datapoints out of the ones that have the \textit{same} labels, where the prediction of kNN and QWEN-7B-Chat match.}
  \label{tab:qwen-same-label-align-stats}
  \vspace{-1mm}
\end{table}

\setlength{\tabcolsep}{7pt}
\renewcommand{\arraystretch}{1.2}
\begin{table}[!t]
\fontsize{10pt}{10pt}\selectfont
  \begin{tabular}{p{24mm}
  >{\centering\arraybackslash}p{11mm}
  >{\centering\arraybackslash}p{11mm}
  >{\centering\arraybackslash}p{11mm}
}
    \toprule
    Datasets & $k=10$ & $k=20$ & $k=30$ \\
    \midrule
    AG News & 85.01 & 89.06 & 89.79 \\
    News Category & 61.78 & 72.55 & 77.23 \\
    AuTexTification  & 59.54 & 59.32 & 61.4 \\
    Implicit Prompts  & 19.84 & 18.82 & 22.88\\
    SST-2 & 64.27 & 68.43 & 70.19 \\
    SemEval-2019  & 43.68 & 45.38 & 45 \\
    \bottomrule
  \end{tabular}
  \caption{Percentage of datapoints out of the ones that have the \textit{different} labels, where the prediction of kNN and QWEN-7B-Chat match.}
  \label{tab:qwen-diff-label-align-stats}
\end{table}

Similar to Table~\ref{tab:gpt-same-label-align-stats} and~\ref{tab:gpt-diff-label-align-stats}, Table~\ref{tab:llama-same-label-align-stats} and Table~\ref{tab:llama-diff-label-align-stats} present results for Llama-3.1-8B under the same scenario. Interestingly, the percentage of data points with different labels of examples in Table~\ref{tab:llama-diff-label-align-stats}  for the labels predicted by kNN and the Llama-3.1-8B shows high agreement when compared to Table~\ref{tab:gpt-diff-label-align-stats}. Further, we extend the analysis to QWEN-7B-Chat and present results in Table~\ref{tab:qwen-same-label-align-stats} and~\ref{tab:qwen-diff-label-align-stats}. In this case as well, we observe a similar behavior as both Gpt-4o-mini and Llama-3.1-8B, indicating that the underlying patterns remain consistent across different settings.

\end{document}